\documentclass[numbers]{mhug}  %
\ifPDFTeX\fi  %
\makeatletter
\patchcmd{\maketitle}{\mhug@titlefont}{\mhug@titlefont\fontsize{15}{18}\selectfont}{}{\typeout{MHUGPATCH: title font patch failed}}
\makeatother

\usepackage[dvipsnames]{xcolor}
\usepackage{soul}

\newcommand{\rg}{ReGAP\xspace}
\newcommand{\bench}{ReliaNeedle\xspace}
\newcommand{\dataset}{EviCe\xspace}
\newcommand{\methodname}{RelyGen\xspace}

\definecolor{boxteal}{HTML}{23776D}   %
\definecolor{boxslate}{HTML}{233A42}  %
\definecolor{boxblue}{HTML}{357CBC}   %
\usepackage[most]{tcolorbox}

\definecolor{boxteal}{HTML}{23776D}
\newtcolorbox{questionbox}{enhanced, breakable=false,
  colback=boxteal!6, colframe=boxteal, boxrule=0.8pt, arc=2mm,
  left=4pt, right=4pt, top=3pt, bottom=3pt, before skip=4pt, after skip=6pt}

\usepackage{microtype}

\usepackage{afterpage}  %
\usepackage{wrapfig}  %
\newsavebox{\sidecapbox}\newsavebox{\sidecapboxb}\newlength{\sidecapw}  %
\usepackage{pgfplots}
\pgfplotsset{compat=1.18}
\usetikzlibrary{patterns}

\usepackage{pifont} %
\usepackage{graphicx} %
\graphicspath{{figs/}} %
\usepackage{xspace} %

\definecolor{cvprblue}{rgb}{0.21,0.49,0.74}
\usepackage{etoc}
\usepackage{tcolorbox}
\tcbuselibrary{skins}
\usepackage{upquote}
\definecolor{promptFrame}{HTML}{3E7CA6}
\definecolor{promptBack}{HTML}{F5F8FB}
\newtcolorbox{promptbox}[1]{enhanced, colback=promptBack, colframe=promptFrame, coltitle=white, fonttitle=\bfseries\footnotesize, title={#1}, boxrule=0.6pt, arc=2pt, left=5pt, right=5pt, top=2pt, bottom=2pt, toptitle=1pt, bottomtitle=1pt, before skip=5pt, after skip=5pt, before upper={\scriptsize\setlength{\topsep}{2pt}\setlength{\partopsep}{0pt}\setlength{\parskip}{0pt}}}
\newcommand{\promptrole}[1]{\par\noindent{\bfseries\color{promptFrame}#1}\par}
\AtBeginDocument{\etocdepthtag.toc{mainmatter}}

\title{When Does Temporal Grounding Help Video Reasoning?}

\author{Yan Shu}
\author[*]{Thomas De Min}
\author[*]{Moreno D'Inc\`a}
\author{Massimiliano Mancini}
\author{Paolo Rota}
\affiliation{University of Trento}
\contribution[*]{Equal contribution}

\abstract{\begingroup\renewenvironment{abstract}{}{}\begin{abstract}
Temporally Grounded Reasoning (TGR) models first predict the video spans that support their answer, and then answer the question. Whether the answer actually depends on these spans is still unclear, and existing benchmarks were not designed to measure this dependence, which we call grounding reliance.  We study this by introducing \bench, a controlled needle-in-a-haystack benchmark whose answers require the visual evidence inside the annotated span. To measure reliance without ground-truth span annotations, we introduce \rg, which compares the confidence change after deleting the predicted span with that after deleting equally long random segments, and validate it on \bench. 
Despite strong results on standard metrics, most TGR models rely on their predicted spans less than their pretrained backbones do.
We trace this weakness to training data and recipes. 
For the data, we introduce \methodname to annotate VideoQA pairs with correctly located, necessary evidence spans; training on just 20k such samples more than doubles ReGAP. Our training analysis shows that supervised initialization is needed to couple answering with localization. Subsequent reinforcement learning with a grounding-only reward improves both reliance and accuracy, whereas rewarding answers erodes reliance.
Finally, better temporal grounding improves accuracy only for high-reliance models, showing that temporal grounding helps video reasoning only when the answer relies on it.
\end{abstract}\endgroup}

\date{\today}
\correspondence{\email{yan.shu@unitn.it}, \email{paolo.rota@unitn.it}}
\metadata[Project page]{\url{https://shuyansy.github.io/TGR-Reliance/}}

\begin{document}

\maketitle

\section{Introduction}
\label{sec:intro}

\begin{figure}[t]
  \centering
  \includegraphics[width=\linewidth]{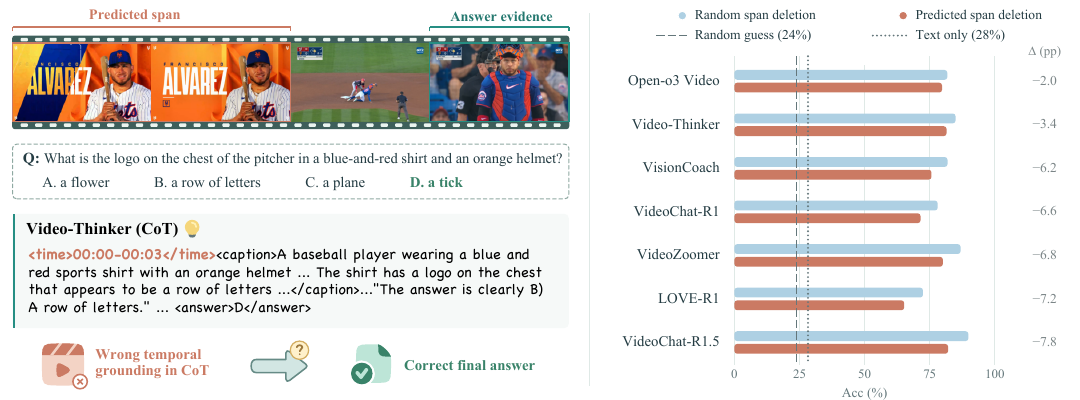}
\caption{\textbf{Grounding without reliance.} (Left) Video-Thinker~\cite{wang2025video} predicts a span that misses the evidence, yet still outputs the correct answer. (Right) Accuracy of open-source TGR models on Video-Oasis~\cite{video-oasis} after deleting each model's predicted span or a random span of equal length, measured on the questions each model answers correctly with the full video. Vertical lines indicate random guessing and text-only accuracy, averaged across models.}
  \label{fig:teaser}
\end{figure}

Answering a question about a video often depends on a few moments in a long stream of frames. Yet, standard video-language models~\cite{zohar2025apollo} locate these moments implicitly, leaving it unclear which visual evidence supports their answers. \emph{Temporally Grounded Reasoning} (TGR) makes this explicit: much like chain-of-thought exposes intermediate steps in language models~\cite{wei2022chain}, TGR asks a model to anchor its reasoning to a specific time span in the video. Recent TGR models improve on tasks such as video question answering (VQA), attributing these results to better visual evidence extraction~\cite{wang2025video, li2025videochat, liu2026videomind, pan2026timesearch, ding2026videozoomer}. However, higher accuracy says little about whether the predicted span drives the answer: a model may miss the span yet answer correctly (\cref{fig:teaser}, left). %
This raises a core question: do models \emph{rely} on their own predicted spans when answering, and to what extent?

As an initial test, we remove models' predicted spans on correctly answered questions from Video-Oasis~\cite{video-oasis} and compare the accuracy drop with deleting random spans of equal length. If answers depend on the predicted evidence, deleting it should be more damaging. This is not what we observe: accuracy remains comparable under both deletions and far above simple chance and the text-only baselines (\cref{fig:teaser}, right). 
Note that these results could reflect weak reliance as well as incorrect grounding. 
Telling the two apart requires knowing where the evidence lies and that the answer depends on it, yet existing benchmarks do not provide this, as deleting their annotated spans barely changes performance (\cref{fig:audit}).
Standard metrics such as accuracy gated by IoU~\cite{nextgqa, cgbench} also fail to establish reliance, since a correct answer paired with a correct span does not imply that the model used that span.

We fill this gap by delving deep into the dependence between reliance and grounding in video reasoning. %
To overcome the limits of existing benchmarks, we introduce the Reliance Gap (\rg), a metric that measures the decrease in the model's confidence in its answer when its predicted span is deleted, relative to deleting a randomly selected segment of equal length. \rg does not require ground-truth annotations and can therefore be applied regardless of their availability or quality. Furthermore, to validate reliance %
in a controlled setting, %
we construct \bench, a needle-style benchmark in which evidence clips are inserted into long, unrelated videos and paired with questions that can only be answered from the inserted clips. %
Surprisingly, when diagnosing TGR models~\cite{lee2026visioncoach,yan2026videochat,li2025videochat,meng2025open,wang2025video,zeng2026video,fu2025love,ding2026videozoomer,meng2026openo3v2,ouyang2026conan} on \bench with \rg, we find that, while they achieve higher accuracy, %
most of them\textit{ rely even less} %
on their predicted spans than their backbone. 

We studied this issue on two axes: the training data and the training method. In TGR training data (\ie, video question--answer pairs annotated with evidence spans), 
these spans are often misplaced or unnecessary for answering, teaching models to predict a span without using it. 
To address this, we introduce \methodname, an automatic pipeline that annotates VideoQA pairs~\cite{zhang2024llavavideo} with temporal evidence spans. Training on these annotations teaches models to rely on their predicted spans.
Its core component, \emph{evidence certification}, keeps a sample only if its span is both sufficient and necessary for the answer. Running \methodname on public VideoQA data yields \dataset, a $20$k-sample evidence-certified training set. 
We show that post-training two existing TGR models on \dataset improves \rg on \bench by more than 2×, whereas training on 20k uncertified samples lowers it.
Examining the training method on this curated data, we further show that SFT followed by GRPO~\cite{shao2024deepseekmath} rewarded solely for temporal grounding improves both reliance and accuracy, whereas skipping SFT leaves reliance low.

Finally, we test whether better localization leads to better answers, by adding grounding data during training and replacing predicted spans with the known evidence at inference. In both cases, low-reliance models gain little accuracy, while models with stronger reliance gain more. %
This confirms that temporal grounding improves video reasoning only when models are trained to rely on their predicted spans.

In summary \ding{202} We study \emph{grounding reliance}, \ie, whether a model's answer depends on its own predicted span. \ding{203} We introduce \rg, an annotation-free metric that measures grounding reliance, and \bench, a controlled benchmark for validating it. Together, they reveal weak reliance in existing TGR models. \ding{204} We analyze how training data and objectives shape reliance, proposing \methodname, a data curation pipeline, that improves both reliance and answer accuracy. \ding{205} We show that improved grounding translates into accuracy gains only when models rely on it, motivating reliance as a target alongside grounding accuracy.

\section{Related Work}
\label{sec:related}

\paragraph{Temporally grounded reasoning.}
Video models that reason step by step in text~\cite{feng2026video, wang2026videorft} can drift away from what the video shows and hallucinate details~\cite{maaz2025video, chen2026think}. Building on video LLMs~\cite{shu2025videoxl} that localize moments in time~\cite{ren2024timechat,liu2026timescope}, TGR ties the reasoning to explicit time spans to keep it faithful to the visual evidence. TGR models fall into two lines of work. \emph{Single-turn} models claim evidence by writing timestamps or frame indices into their reasoning and answer from the original video in one pass. Some are trained with supervised fine-tuning to point to the evidence~\cite{ghazanfari2026chain, gupta2025toga}, but most use reinforcement learning, typically GRPO~\cite{shao2024deepseekmath}. These RL approaches reward either the answer alone~\cite{wang2025video} or also %
the temporal IoU between the predicted and annotated spans~\cite{li2025videochat, yan2026videochat}. Many works refine this recipe with finer grounding rewards~\cite{cheng2026graphthinker, chen2026long, li2026timethink, maaz2025video, xia2026ser}, changes to the training procedure~\cite{meng2025open, zeng2025factorized, lee2026visioncoach}, or new reasoning data~\cite{han2025videoespresso, zhang2026rewatch}.

\emph{Multi-turn} models instead fold agentic video search~\cite{wang2024videoagent, fan2024videoagent, min2024morevqa, wang2025videotree, ma2025drvideo, zhang2026deep, qiu2026longvideo} into a single video LLM that re-watches predicted spans across turns. Most iterate through localizing a segment, re-encoding it at higher resolution or density, and reasoning over it~\cite{xie2025video, li2026video, wang2025videoloop, ouyang2026conan, ma2026act2see}, sometimes with explicit planning or reflection~\cite{liu2026videomind, jain2026sage, li2026videothinker, liu2026videotemp}. They are typically trained with RL after a cold start on distilled search traces, rewarding not only the answer but also tool use~\cite{ding2026videozoomer, he2026framethinker, zhang2026thinking, jain2026sage, zhang2026eva}, temporal overlap with annotated spans~\cite{fu2025love, chen2026think, shen2025zoom, zeng2026video, liu2026videotemp, yang2026longvt}, or evidence sufficiency~\cite{pan2026timesearch, shen2025zoom, chen2026think, zhang2026eva}. Yet several works note that answers can still ignore the retrieved clips~\cite{fu2025love, he2026framethinker, zeng2026video, yang2026longvt}. 

Across both single- and multi-turn models, no existing design or reward ensures that the answer actually depends on the predicted span, a crucial element for trustworthiness.

\paragraph{Evaluating grounded reasoning.}
Grounded VideoQA benchmarks pair each question with annotated evidence and score both the answer and its temporal grounding. Most count an answer as grounded only if the predicted span overlaps the annotated one~\cite{nextgqa, qin2025question, chen2024rextime, cgbench}, while others report the two scores separately or average them~\cite{di2024grounded, cheng2026v}. Recent works extend evaluation to spatial boxes or masks and textual event descriptions~\cite{meng2026videozerobench, wang2026evidence, huang2026eg}, judge-scored reasoning traces~\cite{nagrani2025minerva}, the visual evidence behind implied meaning~\cite{li2026vimu}, or consistency between the reasoning and the answer~\cite{maaz2025video}. Across these benchmarks, however, models answer correctly far more often than they locate the evidence~\cite{nextgqa, chen2024rextime, cgbench, cheng2026v, wang2026evidence, huang2026eg}. Meanwhile, a parallel line audits the questions themselves and finds that many can be answered without the video or its temporal order~\cite{cores2024lost, vbenchcomp, video-oasis}. 
Existing evaluations score agreement with annotations or test whether a question needs the video at all, but none asks whether a model's answer depends on the evidence it claims.

\section{Measuring Reliance}
\label{sec:measuring}

TGR models are evaluated with two metrics: answer accuracy, and the overlap between the predicted span and the annotated one. Each metric scores one output on its own, so neither measures the link between the two. A model can predict the correct span and still take its answer from other frames or from the question alone. Both scores are then high, although the span played no part in the answer.
We define \emph{grounding reliance} (hereafter simply \emph{reliance}) as the extent to which a model's answer depends on its own predicted span, and ask whether TGR models rely on the spans they predict.
To answer this, we first show that existing grounded benchmarks cannot measure reliance and build \bench, a benchmark where the evidence is known and required to answer (\cref{sec:needle}). We then introduce \rg, a metric that measures reliance without ground-truth annotations (\cref{sec:rg}), validate it on \bench (\cref{sec:validate}), and use it to diagnose TGR models (\cref{sec:diag}).

\begin{figure}[t]
  \centering
  \begin{minipage}[b]{0.45\linewidth}\centering
    \includegraphics[width=\linewidth]{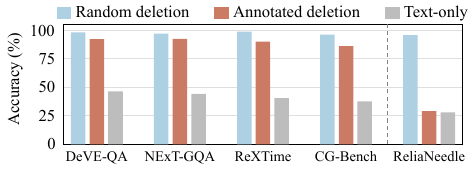}
  \end{minipage}\hfill
  \begin{minipage}[b]{0.52\linewidth}\centering
    \includegraphics[width=\linewidth]{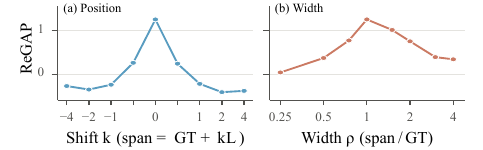}
  \end{minipage}\par
  \begin{minipage}[t]{0.45\linewidth}
 \caption{\textbf{Auditing grounding annotations.} Accuracy of Qwen2.5-VL-7B on its correctly answered questions after deleting a random span or the annotated evidence span. Text-only is the accuracy from the question and options alone, on all questions.}
  \label{fig:audit}
  \end{minipage}\hfill
  \begin{minipage}[t]{0.52\linewidth}
  \caption{\textbf{Validating \rg with controlled spans on \mbox{\bench}.} The predicted span of Qwen2.5-VL-7B is replaced with the ground-truth (GT) span of length $L$, either (a) shifted in position by $kL$ or (b) scaled in width by $\rho$ around its center.}
  \label{fig:validation}
  \end{minipage}
\end{figure}

\subsection{\texorpdfstring{Existing Benchmarks Cannot Isolate Reliance}{Existing Benchmarks Cannot Isolate Reliance}}
\label{sec:needle}
We first investigate whether existing grounding benchmarks can evaluate reliance by checking if their annotated spans are actually required to answer. We study four benchmarks commonly used to evaluate TGR models, DeVE-QA~\cite{qin2025question}, NExT-GQA~\cite{nextgqa}, ReXTime~\cite{chen2024rextime}, and CG-Bench~\cite{cgbench}, which pair each question with an annotated evidence span. Because they score a predicted span by its overlap with this annotation, they implicitly assume that the annotated span holds the evidence the answer requires. On questions a model answers correctly, we compare the model's accuracy when deleting the annotated span with the accuracy when deleting a random segment of the same length. As shown in~\cref{fig:audit}, deleting the annotated span leaves Qwen2.5-VL-7B~\cite{bai2025qwen25vltechnicalreport} at around 90\% accuracy on all four benchmarks, only slightly below the accuracy after a random deletion (other models in \cref{app:audit}).
A correct answer without the annotated span can come from only two sources: \ding{202} \emph{visual leakage}, where the annotated span misses part of the evidence or redundant cues appear elsewhere in the video, or \ding{203} \emph{language priors}, where the question can be answered without video evidence at all. On these benchmarks, text-only accuracy is around $40\%$, well above chance, so language priors alone already answer many questions. Yet the other $50$\% is dominated by visual leakage. Because models can answer correctly without ever seeing the annotated span, existing benchmarks cannot test whether grounded reasoning actually uses localized evidence. Eliminating both confounders requires a benchmark where the evidence is strictly confined to a known span and indispensable for answering. %
A needle-in-a-haystack design meets both requirements by construction.
However, such existing benchmarks~\cite{videoniah, mlvu, nemo}
either insert synthetic needles or insert real clips that the question locates by describing their content or giving their timestamp, and do not check for leakage or language priors. 
None captures the real-event reasoning of TGR models.

To address the issues of previous benchmarks, \textbf{we build} \textbf{\bench}. We insert a real action clip (the needle) into an unrelated video and write a multiple-choice question that can be answered only from the needle. Human annotators then check each sample, confirming that the question is well posed and answerable only from the needle, and that the needle shares no content with the haystack.
\bench comprises 725 videos, each with one question and the annotated needle spans (details in \cref{app:needle}). The \bench{} group in \cref{fig:audit} confirms this design, as removing the needle reduces Qwen2.5-VL-7B’s accuracy to below $30\%$, near chance ($25\%$), compared to $~90\%$ on existing benchmarks.

\subsection{Reliance Gap (ReGAP)}
\label{sec:rg}
To measure reliance, we cannot simply use accuracy and span overlap: %
even a correct answer paired with a well-grounded span does not establish that the 
model used that span. %
Intuitively, however, a model that truly \textit{relies} on its predicted temporal grounding span to answer a question should become less confident when that span is removed from the video. %
Based on this principle, we propose Reliance Gap (\rg), an annotation-free metric that quantifies reliance by intervening on the model's own predicted span. %

Formally, given a video $v$ and a question $q$ with answer options $\mathcal{A}=\{a_1,\cdots,a_n\}$, the model outputs a predicted span
$\hat{s}\subseteq v$ and an answer $\hat{a}\in\mathcal{A}$\footnote{We discuss open-ended answers in \cref{app:rg}.}. Let $p(v)=\Pr(\hat{a}\mid v, q, \mathcal{A})$ be the model's probability for $\hat{a}$, renormalized over $\mathcal{A}$, when shown the video $v$ with question $q$. We delete the predicted span and, as a control, $K$ random contiguous blocks $r_k$ of the same length of the predicted span from the rest of the video. Each block is deleted separately, while keeping $\hat{a}$, $q$, and $\mathcal{A}$ fixed. We then compute:
\begin{equation}
c = \operatorname{logit}\big(\mathbb{E}_k[p(v\setminus r_k)]\big) - \operatorname{logit}\big(p(v\setminus\hat{s})\big),
\end{equation}
where $\mathbb{E}_k$ averages over the $K$ blocks, $\operatorname{logit}(p)=\log\frac{p}{1-p}$.
Here, $v \setminus r_k$ denotes $v$ with block $r_k$ removed, and $v \setminus \hat{s}$ denotes $v$ with the predicted span removed. A positive $c$ means deleting inside the predicted span costs the model more than deleting an equal amount elsewhere, that is, the predicted span carries evidence the answer depends on.

Since models differ in how confident they are, we normalize by each model's confidence and define \rg as:
\begin{equation}
\operatorname{\rg} \;=\; \frac{\mathbb{E}[c]}{\tau}, \qquad \tau \;=\; \mathbb{E}\big[\,\big|\operatorname{logit}(p(v))\big|\,\big],
\end{equation}
where expectations run over the evaluation items, and $\tau$ is computed separately for each model.

\rg is unbounded, and its sign shows whether the predicted span matters more than random frames. Intuitively, \rg $= 0$ means the predicted span plays no special role in answering, since deleting it affects the model no more than deleting random frames (\eg, when answering from language priors or global context). \rg $< 0$ means the predicted span matters even less than random frames, because the model draws its answer from visual cues outside it, so deleting $\hat{s}$ leaves those cues intact while random deletions elsewhere occasionally remove the frames it used (details in \cref{app:rg}).
At \rg $= 1$, deleting the predicted span typically lowers the model's probability for its answer to about $0.5$, showing that its prediction rests on those frames. Values beyond $1$ indicate even stronger reliance, as deleting the span pushes the model toward other answers.

\subsection{Validating ReGAP}
\label{sec:validate}
\begin{lrbox}{\sidecapbox}\scriptsize\setlength{\tabcolsep}{2pt}%
  \begin{tabular}{lccccc}
    \toprule
    & & & \multicolumn{3}{c}{\rg} \\
    \cmidrule(lr){4-6}
    Model & Acc. & R@0.3 (\%) & All & Hit & Miss \\
    \midrule
    \multicolumn{6}{l}{\emph{General-purpose}} \\
    Qwen2.5-VL-7B~\cite{bai2025qwen25vltechnicalreport} & 57.2 & 46.0 & 0.598 & 0.945 & 0.302 \\
    Qwen3-VL-4B$^\dagger$~\cite{bai2025qwen3} & 57.4 & 55.5 & 0.711 & 1.026 & 0.320 \\
    \midrule
    \multicolumn{6}{l}{\emph{TGR, single-turn}} \\
    VisionCoach~\cite{lee2026visioncoach} & 65.8 & 62.3 & 0.468 & 0.763 & $-$0.019 \\
    VideoChat-R1.5~\cite{yan2026videochat} & 61.7 & 47.1 & 0.438 & 0.863 & 0.060 \\
    VideoChat-R1~\cite{li2025videochat} & 52.6 & 46.8 & 0.375 & 0.709 & 0.081 \\
    Open-o3-Video~\cite{meng2025open} & 66.8 & 42.9 & 0.271 & 0.764 & $-$0.100 \\
    Video-Thinker~\cite{wang2025video} & 54.1 & 17.5 & $-$0.011 & 0.384 & $-$0.095 \\
    \midrule
    \multicolumn{6}{l}{\emph{TGR, multi-turn}} \\
    Video-o3~\cite{zeng2026video} & 67.3 & 64.7 & 0.611 & 0.700 & 0.448 \\
    LOVE-R1~\cite{fu2025love} & 58.5 & 67.1 & 0.493 & 0.628 & 0.218 \\
    Conan~\cite{ouyang2026conan} & 60.0 & 19.6 & 0.240 & 0.489 & 0.180 \\
    VideoZoomer~\cite{ding2026videozoomer} & 64.3 & 7.7 & 0.195 & 0.693 & 0.153 \\
    Open-o3-Video-v2$^\dagger$~\cite{meng2026openo3v2} & 65.5 & 77.4 & 0.597 & 0.651 & 0.413 \\
    \bottomrule
  \end{tabular}
\end{lrbox}%
\begin{wraptable}[22]{r}{\wd\sidecapbox}
\centering
\vspace{-\baselineskip}
\usebox{\sidecapbox}
\caption{\textbf{Diagnosis on \bench.} \textbf{R@0.3~(\%)} is the percentage of predicted spans that overlap the ground-truth evidence with $\text{IoU} \ge 0.3$. \rg is reported across all evaluation items (\textbf{All}) and separately on items where the predicted span hits ($\text{IoU} \ge 0.3$, \textbf{Hit}) or misses ($\text{IoU} < 0.3$, \textbf{Miss}) the evidence. $^\dagger$ marks Qwen3-VL-4B and the TGR model built on it; all other TGR models are built on Qwen2.5-VL-7B.}
\label{tab:diag}
\end{wraptable}
A valid reliance metric should score high when a predicted span captures the frames a model uses to answer, and low when it does not. We test this on \bench, where all evidence required to answer is confined to the ground-truth needle of length $L$ by construction. Holding Qwen2.5-VL-7B's full-video answer fixed, we replace its predicted span $\hat{s}$ with controlled spans shifted or scaled around the needle to set how much of the evidence is deleted, and track \rg.

\noindent\textbf{Position.} Shifting the span by $kL$ keeps its length fixed while moving it away from the needle, testing whether \rg tracks where the used frames lie rather than deletion length alone. In \cref{fig:validation}a, \rg peaks above $1$ at the needle ($k=0$), drops as overlap decreases, and turns negative once the span covers only unrelated background ($|k|\ge 1$). 

\noindent\textbf{Width.} Scaling the span by $\rho$ around the needle center keeps it on the evidence while varying its width, testing whether \rg can be inflated by spans that are too narrow or too wide. In \cref{fig:validation}b, \rg peaks at $\rho=1$, where the span matches the needle. Narrower spans ($\rho<1$) delete part of the needle and leave the rest for the model to use, whereas wider spans ($\rho>1$) add unrelated frames without removing further evidence. 
The random deletions, which match the span length, grow with it and remove more context, narrowing the gap between the two. Together, these results show that, without access to ground-truth annotations, \rg rewards spans that tightly isolate the frames a model uses.

\begin{figure*}[tp]
  \centering
  \includegraphics[width=\textwidth, trim=4 5 6 4, clip]{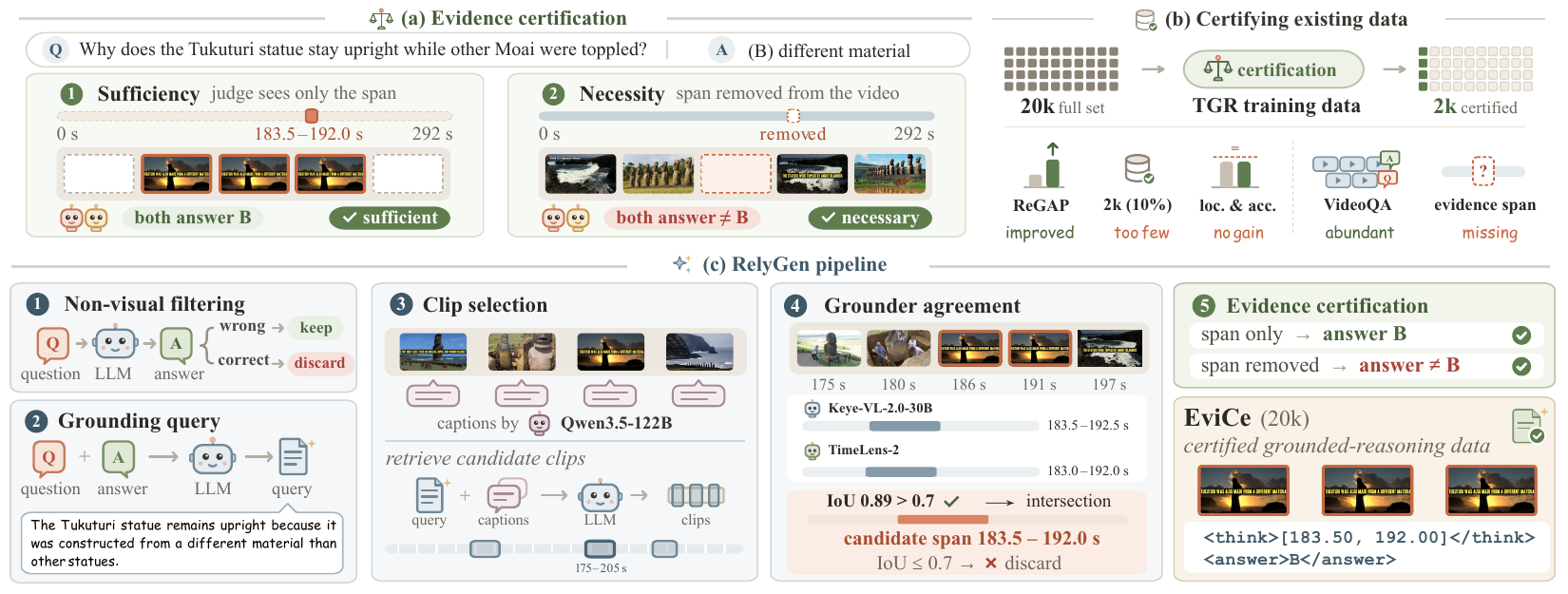}
  \caption{\textbf{Evidence certification and \methodname.} (a) A sample is kept only if two judges answer correctly from its span alone and fail when the span is deleted. (b) Certifying TGR training data improves \rg (\cref{tab:pilot}), but the $2$k certified samples are too few to improve localization and accuracy, while VideoQA data is abundant but lacks evidence spans. (c) \methodname turns VideoQA pairs without span annotations into candidate spans and certifies them with the same tests to build \dataset.%
}
  \label{fig:pipeline}
\end{figure*}

\subsection{Reliance in Existing TGR Models}
\label{sec:diag}

We use \rg to diagnose ten open TGR models, five single-turn models that predict a span and an answer in one pass, and five multi-turn agents that crop and re-watch the video over several turns before answering (\cref{tab:diag}). Nine are built on Qwen2.5-VL-7B and one on Qwen3-VL-4B, and each is compared with its own backbone. For this experiment we use \bench, where the known evidence location allows us to split each model's reliance by whether its predicted span hits the evidence. Each model runs with its own prompt and inference procedure (\cref{app:diag}). From \Cref{tab:diag}, we draw three takeaways: %

\noindent\textbf{TGR models gain more in answer accuracy than in temporal grounding}. Most TGR models achieve higher accuracy than the backbone, but only some increase the fraction of predicted spans that hit the ground-truth evidence ($\text{IoU} \ge 0.3$, \textbf{R@0.3}), and a few hit it far less often.  

  \noindent\textbf{Better grounding does not come with higher reliance.} Even TGR models that ground the evidence better than their backbone rely less on their predicted spans overall (\textbf{All}). The only exception is Video-o3, which slightly surpasses the backbone in \rg.

\noindent\textbf{Where does lower reliance come from?} \rg does not require ground-truths but in this analysis we use the grounding annotations of \bench, and split \rg into \textbf{Hit} ($\text{IoU} \ge 0.3$) and \textbf{Miss} ($\text{IoU} < 0.3$) subsets. On \textbf{Miss} items, where the predicted span misses the evidence, single-turn TGR models barely rely on it ($\rg \le 0.081$), whereas the backbone partly does ($0.302$), showing that their wrong spans no longer reflect where the answer comes from; multi-turn models, which re-watch the predicted span, retain more reliance, although only Video-o3 and Open-o3-Video-v2 exceed their backbones. On \textbf{Hit} items, every TGR model achieves a lower \rg than its backbone, 
meaning that even when TGR models localize the right segment, they rely on it less to answer.

\section{\texorpdfstring{Building Reliance via Data and Training}{Building Reliance via Data and Training}}
\label{sec:why}
\Cref{sec:diag} shows that TGR models have lower reliance than their backbone. 
We first investigate whether the low reliance of TGR models stems from their training data.

\paragraph{Low reliance originates from TGR training data.} 
We consider the TGR models of \cref{sec:diag}, and apply the deletion experiment of \cref{sec:needle} to their training data, namely NExT-GQA~\cite{nextgqa}, CG-Bench~\cite{cgbench}, Video-Thinker-10K~\cite{wang2025video}, and Seeker-173K~\cite{zeng2026video}.
As in \cref{sec:needle}, deleting the annotated spans leaves accuracy high ($81.9$--$90.8\%$) and comparable to random deletion (full results in \cref{app:audit_train}).
Thus, training on samples with noisy annotations rewards predicting the annotated span while allowing the answer to come from other frames or from the question alone, teaching the model to predict a span without actually relying on it.

\paragraph{Can TGR training data be cleaned?} 
To clean such data, we aim to retain samples whose visual evidence is \textit{sufficient} and \textit{necessary} to answer, and build a two stage certification method (\cref{fig:pipeline}a). First, \textit{sufficiency} is tested by showing a multimodal judge only the annotated span and checking that it answers correctly, filtering out spans that lack visual evidence. Second, we impose \emph{necessity} by deleting the annotated span from the full video and requiring the judge to answer \textit{incorrectly}, thereby filtering out samples whose answer leaks from surrounding frames or can be solved from the question alone. These two steps retain only samples in which the answer is contained in the annotated span and cannot be determined without it. We call this procedure \emph{evidence certification}. 
To mitigate biases, we require agreement between two multimodal judges, neither of which is among the models we train or evaluate (\cref{app:certification}).

\begin{wraptable}[10]{r}{0.46\linewidth}
  \centering
\vspace{-\baselineskip}
  \footnotesize
  \setlength{\tabcolsep}{4pt}
  \begin{tabular}{@{}lcccc@{}}
    \toprule
    Model & Size & Acc. & R@0.3 & \rg \\
    \midrule
    Qwen2.5-VL-7B~\cite{bai2025qwen25vltechnicalreport} & -- & 57.2 & 46.0 & 0.60 \\
    \quad + TGR-Mix & 20k & 58.2 & 61.3 & 0.44 \\
    \quad + TGR-Mix, random & 2k & 57.0 & 46.6 & 0.48 \\
    \quad + TGR-Mix, certified & 2k & 57.2 & 46.8 & \textbf{0.75} \\
    \bottomrule
  \end{tabular}
  \caption{\textbf{Certifying TGR training data.} Fine-tuning on TGR-Mix, on the $2$k samples of TGR-Mix that pass evidence certification, or on a random $2$k subset of TGR-Mix.}
  \label{tab:pilot}
\end{wraptable}
\paragraph{Clean TGR data improves reliance.} 
We apply evidence certification to \emph{TGR-Mix}, a $20$k-sample mixture of the training sets audited above (\eg, NExT-GQA and CG-Bench).
Only $2$k of these samples, about $10\%$, pass.
We then fine-tune Qwen2.5-VL-7B on the full TGR-Mix, on the $2$k certified samples, and on $2$k samples drawn at random from TGR-Mix, which tells us whether a change in reliance is simply due to training on less data.
\Cref{tab:pilot} shows that training on the full TGR-Mix improves grounding (R@0.3) on \bench but lowers \rg from $0.60$ to $0.44$, which is the same pattern we observed in \S\ref{sec:diag}.
The random $2$k subset lowers \rg as well (to $0.48$), and only the certified subset raises it, to $0.75$.
Because a random subset of the same size does not help, we attribute the improvement to certification.

\subsection{\texorpdfstring{Data curation pipeline (\methodname)}{Data curation pipeline (RelyGen)}}
\label{sec:relygen}
We now scale up certified data. 
Certification keeps only $10\%$ of existing TGR training data, which is enough to raise \rg but too little to improve grounding and accuracy (\cref{tab:pilot}), so we need data from another source.
VideoQA data (\eg, LLaVA-Video-178K~\cite{zhang2024llavavideo}) is abundant but has no grounding annotations. We therefore build an automatic annotation pipeline, \methodname (\cref{fig:pipeline}c).
First, an LLM removes the questions that can be answered without visual input, \ie, text-only (\ding{202}). For the remaining questions, we look for the clips that contain the evidence. An LLM rewrites each question and answer pair into a description of the target visual event, which we call the \emph{grounding query} (\ding{203}). We then caption each video in short clips and retrieve the clips whose captions match the grounding query (\ding{204}). Two different off-the-shelf models ground the query within each retrieved clip, and we keep a candidate span only when both agree (\ding{205}). Finally, evidence certification keeps only the candidate spans that are sufficient and necessary for the answer (\ding{206}).

We use \methodname to build \dataset, a training set of $20$k evidence-certified samples. In each sample, the annotated span matches where the queried event takes place and is needed to answer the question. We provide full \methodname details and \dataset statistics in \cref{app:pipeline}.

\begin{table}[t]
  \centering
  \scriptsize
  \setlength{\tabcolsep}{3pt}
  \begin{tabular*}{\linewidth}{@{\extracolsep{\fill}}lcccccc}
    \toprule
    & & \multicolumn{3}{c}{\bench} & \multicolumn{2}{c}{Video-Oasis} \\
    \cmidrule(lr){3-5}\cmidrule(lr){6-7}
    Model & Size & Acc. & R@0.3 & \rg & Acc. & \rg \\
    \midrule
    Qwen2.5-VL-7B~\cite{bai2025qwen25vltechnicalreport} & --   & 57.2 & 46.0 & 0.60 & 29.8 & 0.09 \\
    \midrule
    \multicolumn{7}{l}{\emph{(a) Fine-tuning the backbone}} \\
    \quad + ReXTime~\cite{chen2024rextime}      & 10k  & 57.0 & 54.9 & 0.45 & 28.2 & 0.07 \\
    \quad + Video-Thinker~\cite{wang2025video} & 10k  & 58.3 & 58.3 & 0.40 & 30.0 & 0.06 \\
    \quad + \methodname{} w/o cert.          & 20k & 57.9 & 60.0 & 0.42 & 30.5 & 0.07 \\
    \quad + \methodname{} (\dataset)                       & 10k  & 59.0 & 65.2 & 0.85 & 31.6 & 0.34 \\
    \quad + \methodname{} (\dataset)                       & 20k  & \textbf{59.4} & \textbf{68.1} & \textbf{1.07} & \textbf{34.2} & \textbf{0.41} \\
    \midrule
    \multicolumn{7}{l}{\emph{(b) Post-training of TGR models}} \\
    VideoChat-R1.5~\cite{yan2026videochat}    & --  & 61.7 & 47.1 & 0.44 & 29.0 & 0.08 \\
    \quad + \methodname{} w/o cert.                  & 20k & 55.1 & 60.3 & 0.38 & 28.5 & 0.04 \\
    \quad + \methodname{} (\dataset)                      & 20k & \textbf{62.6} & \textbf{68.9} & \textbf{0.99} & \textbf{30.6} & \textbf{0.30} \\
    LOVE-R1~\cite{fu2025love}                 & --   & 58.5 & 67.1 & 0.49 & 30.8 & 0.36 \\
    \quad + \methodname{} w/o cert. & 20k                  & 55.3 & 72.5 & 0.32 & 29.6 & 0.28  \\
    \quad + \methodname{} (\dataset)                      & 20k     & \textbf{62.1} & \textbf{74.9} & \textbf{1.20} & \textbf{32.6} & \textbf{0.52}  \\
    \bottomrule
  \end{tabular*}
  \caption{\textbf{Effect of the training data.} (a) Fine-tuning on ReXTime, Video-Thinker, \methodname{} w/o certification, and \dataset in the single-turn format. (b) Existing TGR models further trained in their own formats. Bold marks the best result per model.}
  \label{tab:data}
\end{table}

\subsection{\texorpdfstring{Experimental Setup}{Experimental Setup}}
\label{sec:impl}
Having built \dataset, we study reliance along two axes. We first test whether \dataset teaches reliance better than existing TGR training data (\cref{sec:effect_data}), and then, with the data fixed, which training method best preserves it (\cref{sec:objective}).

We train two backbones, Qwen2.5-VL-7B and Qwen3-VL-8B~\cite{bai2025qwen3} (results in \cref{app:extra}), in two reasoning formats. In the \textit{single-turn} format, the model first predicts the evidence span as timestamps and then answers, \eg, \texttt{<think>[27.0, 50.0]</think> <answer>B</answer>}. In the \textit{multi-turn} format, the model first crops the video via tool calling and then answers after viewing the cropped clip. We evaluate on \bench and on Video-Oasis~\cite{video-oasis}, which has 11,332 questions from 14 established VideoQA benchmarks (\eg, Video-MME~\cite{fu2025videomme}, MVBench~\cite{li2024mvbench}, EgoSchema~\cite{mangalam2023egoschema}, MLVU~\cite{mlvu}) and removes items that are solvable without visual and temporal content. Training details are in \cref{app:training,app:agent}.

\subsection{\texorpdfstring{Training on \dataset}{Training on \dataset}}
\label{sec:effect_data}
We first evaluate whether \dataset can improve reliance and downstream performance. Then, we test whether \dataset can improve the performance of existing TGR models.

\paragraph{\dataset improves reliance and downstream performance.} 
We compare fine-tuning on a random $10$k-sample subset of \dataset{} with fine-tuning on $10$k samples from ReXTime~\cite{chen2024rextime} and Video-Thinker~\cite{wang2025video}. 
\Cref{tab:data}a shows that training on ReXTime and Video-Thinker improves temporal grounding (\eg, $+8.9\%$ R@0.3 for ReXTime) but degrades \rg, lowering it below the backbone's performance for both \bench and Video-Oasis (\eg, $-0.20$ for Video-Thinker in \bench). Training on \dataset improves grounding, accuracy, and reliance on both benchmarks, more than tripling \rg on Video-Oasis ($0.34$ vs $0.09$) and outperforming training on ReXTime and Video-Thinker in all metrics (\eg, $+6.9\%$ R@0.3 on \bench over Video-Thinker). 
These results confirm that training on TGR training data is noisy: while models might ground more accurately compared to the backbone, they become less reliant on those spans when answering, limiting the benefit of improved temporal grounding.
The certified examples in \dataset strengthen this reliance, enabling models to better use grounded evidence and achieve larger gains in both temporal grounding and accuracy.

We then scale training to the full $20$k samples of \dataset.
Doubling the data improves every metric on both benchmarks, \ie, accuracy rises by $0.4\%$ on \bench and $2.6\%$ on Video-Oasis, R@0.3 by $2.9\%$, and \rg by $0.22$ and $0.07$, respectively.
The margin over both datasets therefore grows with data size, \eg, from $+6.9\%$ to $+9.8\%$ R@0.3 over Video-Thinker and from $+0.40$ to $+0.62$ \rg over ReXTime on \bench.
However, the extra data helps only when it is certified, as the same $20$k \methodname samples without certification reach only $0.42$ and $0.07$ \rg, close to these two datasets and well below \dataset-10k.

\paragraph{\dataset improves existing TGR models.}
We post-train two existing TGR models, VideoChat-R1.5~\cite{yan2026videochat} and LOVE-R1~\cite{fu2025love}, following their own formats, on $20$k \dataset samples (\cref{tab:data}b). On both models, \dataset more than doubles \rg on \bench, and improves it on Video-Oasis (\eg, $+0.22$ on VideoChat-R1.5). Moreover, both temporal grounding and accuracy are improved. Conversely, post-training on the same $20$k \methodname samples without certification raises R@0.3 but lowers both \rg and accuracy on both models (\eg, \rg drops from $0.49$ to $0.32$ on LOVE-R1), reproducing the trend of TGR training data. These results show that existing TGR models can benefit from better reliance, improving both their temporal grounding and accuracy, further validating \dataset and \methodname. 

\subsection{\texorpdfstring{Choosing the Training Recipe}{Choosing the Training Recipe}}
\label{sec:objective}
\begin{figure*}[t]
  \centering
  \includegraphics[width=\textwidth]{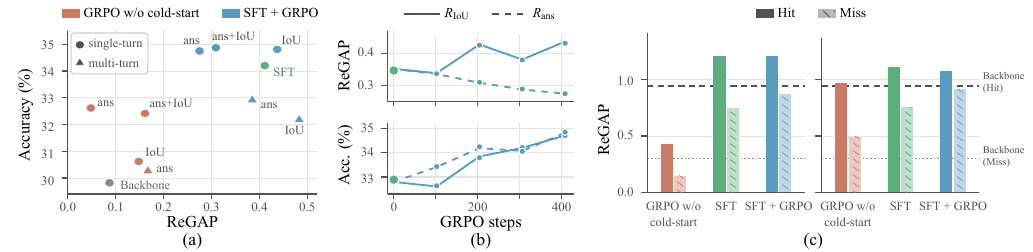}

  \caption{\textbf{Effect of the training recipe.} (a) Accuracy and \rg on Video-Oasis for single-turn and multi-turn models, where labels on GRPO variants denote the reward function. (b) Training dynamics on Video-Oasis for single-turn \emph{SFT + GRPO} under $R_{\text{IoU}}$ or $R_{\text{ans}}$. (c) \rg on \bench when the predicted span hits ($\text{IoU} \ge 0.3$) or misses the evidence across recipes for single-turn (left) and multi-turn (right) models (\emph{GRPO w/o cold-start} uses $R_{\text{ans}}$; \emph{SFT + GRPO} uses $R_{\text{IoU}}$).}
  \label{fig:plane}
\end{figure*}
\label{sec:arch}
With \dataset fixed, we ask how training shape reliance. We compare three paradigms: GRPO~\cite{shao2024deepseekmath} (\emph{GRPO w/o cold-start}) as in most TGR models, supervised fine-tuning (\emph{SFT}), and SFT followed by GRPO (\emph{SFT + GRPO}). 
In \emph{SFT + GRPO}, the data is split, so SFT runs on $1/3$ of \dataset and GRPO on the remaining $2/3$. For GRPO, we compare three rewards: answer accuracy ($R_{\text{ans}}$), answer accuracy and temporal IoU 
($R_{\text{ans+IoU}}$, standard in TGR), and grounding IoU ($R_{\text{IoU}}$), see \cref{app:training}.
Each recipe is trained on both backbones and formats (\cref{sec:impl});
for single-turn Qwen2.5-VL-7B, \emph{SFT} corresponds to the $20$k \dataset model in \cref{tab:data}a. \Cref{fig:plane} shows the results on this backbone, while Qwen3-VL-8B results are in \cref{app:extra}. %

\paragraph{Initialization.} \emph{GRPO w/o cold-start} with $R_\text{ans}$ raises accuracy on Video-Oasis, and adding an IoU term ($R_{\text{ans+IoU}}$ or $R_{\text{IoU}}$) gives a higher \rg than $R_{\text{ans}}$ alone.
Yet its \rg remains far below that of \emph{SFT} (\cref{fig:plane}a). Without SFT to couple answering with grounding, RL treats the predicted span as a detached output, either reinforcing existing answer shortcuts or optimizing grounding in isolation. Splitting \rg into \textbf{Hit} ($\text{IoU} \ge 0.3$) and \textbf{Miss} ($\text{IoU} < 0.3$) subsets of \bench (\cref{fig:plane}c) confirms this. 
On \textbf{Hit} items, models trained solely with \emph{GRPO} rely on their spans no more than the backbone does.
Instead, \emph{SFT} on \dataset increases reliance on grounded evidence: both single- and multi-turn models rely more on both \textbf{Hit} and \textbf{Miss} spans than the backbone, suggesting their answers follow the predicted spans.

\paragraph{Reward design.} We run GRPO from the same cold-start SFT checkpoint with $R_{\text{ans}}$ and with $R_{\text{IoU}}$ (\cref{fig:plane}b). Accuracy improves by a similar amount in both cases, but \rg decreases throughout training with $R_{\text{ans}}$ and increases with $R_{\text{IoU}}$. We believe this happens because an answer reward ($R_{\text{ans}}$ or $R_{\text{ans+IoU}}$) is also given to rollouts that answer correctly with a span that misses the evidence, so the model learns that it can answer without using its span. We therefore recommend \emph{SFT + GRPO} with $R_{\text{IoU}}$ for training TGR models that rely on their spans. The multi-turn model behaves in the same way. Its \rg is higher with $R_{\text{IoU}}$ than with $R_{\text{ans}}$, and with $R_{\text{IoU}}$ it is close to that of the single-turn model.

\section{Grounding Helps Only When the Answer Relies on It}
\label{sec:help}

TGR models are built on the idea that grounding the evidence leads to better answers. We expect this effect to be strongest when answers rely on the predicted span and weaker when the model can answer from other frames or the question alone.
With models trained on \dataset (\cref{sec:effect_data}), which rely on their spans, we can test whether better grounding leads to better answers.
We do so at training time, by adding temporal grounding data and checking whether QA accuracy rises, and at inference time, by correcting the predicted span and checking whether wrong answers are fixed.

\begin{figure*}[t]
  \centering
  \includegraphics[width=\textwidth]{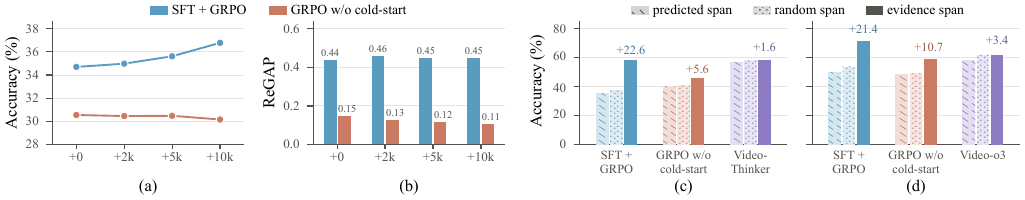}
  \caption{\textbf{Grounding helps only when the answer relies on it.} (a,~b) Scaling grounding data in training. Accuracy and \rg on Video-Oasis as temporal grounding data is added to single-turn GRPO with $R_{\text{IoU}}$, for \emph{SFT + GRPO} and \emph{GRPO w/o cold-start}. (c,~d) Correcting localization at inference on \bench. Accuracy on items where the predicted spans of all three models in a panel miss the evidence ($\text{IoU} < 0.3$; $128$ single-turn items in (c) and $103$ multi-turn items in (d)), with the predicted span kept or replaced by a random span of the same length or by the ground-truth evidence span. Results are averaged over three runs.}
  \label{fig:help}
\end{figure*}

\paragraph{Scaling temporal grounding data during training.}
We first test whether adding pure temporal grounding data to GRPO improves QA accuracy. Unlike the VideoQA samples of \dataset, these samples only require localizing a described event, and they come from a curated external dataset that \methodname never processes. We use TimeLens2-93K~\cite{zhu2026timelens2}, with videos from under a minute to over an hour, and rephrase each grounding query as a question about the described event.
On Qwen2.5-VL-7B, we mix increasing amounts of these samples (up to $10$k),
comparing our recipe (\emph{SFT + GRPO}) against \emph{GRPO w/o cold-start}. 
We inject data from TimeLens2-93K only during GRPO, rewarding it exclusively with $R_{\text{IoU}}$.
Because the answer is never rewarded, extra grounding data can improve QA accuracy only if the model relies on its predicted span to reason. 
Figures~\ref{fig:help}~(a) and~(b) confirm this on Video-Oasis. Since cold-start SFT on \dataset has already established this reliance, scaling grounding data in \emph{SFT + GRPO} lets the extra grounding signal carry over to answering, steadily raising QA accuracy (from $34.7\%$ to $36.8\%$) while keeping reliance high (\rg between $0.44$ and $0.46$). By contrast, because the original model lacks such reliance, optimizing $R_{\text{IoU}}$ in \emph{GRPO w/o cold-start} supervises localization in isolation from answering, yielding no accuracy gain and reducing \rg from $0.15$ to $0.11$.

\paragraph{Correcting temporal grounding at inference.}
We now test whether intervening on temporal grounding at inference time repairs wrong answers on \bench. Focusing on samples where the predicted spans miss the evidence ($\text{IoU} < 0.3$), we replace the grounded span with either the ground-truth or a random one of the same length, and let generation continue from that point. This corresponds to editing the timestamp in the reasoning trace for single-turn models and changing the cropped segment for multi-turn models. 
Replacing the predicted span with the ground-truth one raises the accuracy of \emph{SFT + GRPO} by $22.6$ points in single-turn and $21.4$ in multi-turn models (Figures~\ref{fig:help}~(c) and (d)). In contrast, Video-Thinker ($+1.6$) and Video-o3 ($+3.4$) barely change when given the ground-truth evidence, similar to when their predicted span is replaced by a random one.
\emph{GRPO w/o cold-start}, which relies less on its spans, falls in between ($+5.6$ and $+10.7$). Grounding therefore improves reasoning only when the answer genuinely relies on it.

\section{Conclusion}
\label{sec:conclusion}

In this work, we investigate whether TGR models rely on their predicted spans when answering questions. We show that existing benchmarks cannot measure this, because models keep about $90\%$ of their accuracy even after the annotated evidence is deleted. With the proposed \rg and \bench, we find that nine out of ten TGR models rely on their predicted spans less than their backbone, even when the span is correct. We traced this to the training data and to the training pipeline. The spans in existing training data are not needed to answer the question, so more of this data improves localization but lowers reliance. Training on the $20$k certified samples of \dataset raises \rg above the backbone and improves accuracy, also for existing TGR models. Then, we show that training with SFT followed by GRPO leads to the best reliance. To this extent, rewarding only the predicted grounding obtains the best accuracy and \rg. Finally, better grounding helps mainly the models that rely on it. Replacing the predicted span with the correct one at inference adds more than 20 accuracy points to our models and fewer than 4 to existing TGR models. In conclusion, temporal grounding helps video reasoning only when the answer relies on it, so future TGR models should be evaluated and trained for reliance, alongside accuracy and grounding.

\section*{Acknowledgements}
This work is supported by the EU project ELLIOT (101214398). We acknowledge the CINECA award under the ISCRA initiative for the availability of high-performance computing resources and support. We thank Mountain Map S.r.l.\ for its support of this work. We also thank the Multimedia and Human Understanding Group (MHUG) and Fondazione Bruno Kessler (FBK) for their valuable feedback and insightful suggestions. Thomas De Min is funded by NextGeneration EU.

\clearpage
{
    \small
    \setlength{\bibsep}{0.5pt plus 0.3pt}
    \bibliographystyle{ieeenat_fullname}
    \bibliography{main}
}

\beginappendix
\crefalias{section}{appendix}
\renewcommand{\dblfloatpagefraction}{0.8}  %
\nocite{star,sigurdsson2016hollywood,liu2024etbench,zhao2019hacs,wang2025nextevent,chen2025longvila,farre2024finevideo,grauman2022ego4d,wang2024internvid,huang2024vtimellm,wang2024cosmo,lei2021qvhighlights,hendricks2017didemo,oncescu2021queryd,zala2023hirest,qwen35,kwai2026keyevl2,wang2025internvl35}
\crefalias{subsection}{subappendix}
\crefalias{subsubsection}{subsubappendix}

\etocdepthtag.toc{appendix}
\etocsettagdepth{mainmatter}{none}
\etocsettagdepth{appendix}{subsection}
\etocsettocstyle{\par\vspace{5pt}\noindent{\bfseries Contents}\par\vspace{3pt}}{\par\vspace{4pt}}
\etocsetstyle{section}{}{\par\addvspace{2.5pt}}{\noindent\makebox[1.75em][l]{\makebox[0.75em][r]{\bfseries\etocnumber}}{\bfseries\etocname}\nobreak\leaders\hbox to 0.6em{\hss.\hss}\hfill\nobreak\makebox[1.3em][r]{\etocpage}\par}{}
\etocsetstyle{subsection}{}{\par}{\noindent\hspace*{1.2em}\makebox[2.3em][l]{\makebox[1.3em][r]{\etocnumber}}\etocname\nobreak\leaders\hbox to 0.6em{\hss.\hss}\hfill\nobreak\makebox[1.3em][r]{\etocpage}\par}{}
\tableofcontents

\section{Reliance Gap Protocol}
\label{app:rg}

\paragraph{What \rg measures.} \rg intervenes on the video content inside the predicted span, so it measures whether the span covers the evidence the answer depends on, that is, how faithfully the span explains the answer in the sense of Jacovi and Goldberg~\cite{jacovi2020towards}. It does not by itself test whether producing the span changes the answer, which the interventions on the predicted span in \cref{fig:help}~(c) and (d) test directly. \rg follows deletion-based faithfulness measures, such as comprehensiveness in ERASER~\cite{deyoung2020eraser}, the deletion curve of RISE~\cite{petsiuk2018rise}, and ROAR~\cite{hooker2019benchmark}, as well as the chain-of-thought faithfulness tests of Lanham et al.~\cite{lanham2023measuring} and Turpin et al.~\cite{turpin2023language}. It adapts them to temporal grounding in three ways. It deletes the span the model itself cites rather than regions of an attribution map, it compares this deletion with length-matched random deletions to separate the evidence from the effect of removing frames, and it normalizes by the model's confidence so that models of different confidence can be compared.

\paragraph{Deletion.} Deletion removes the sampled frames inside the span together with a 1\,s margin on each side, and the remaining frames keep their original timestamps. We use $K{=}3$ random blocks per item, each contiguous and deleted separately. Each random block is contiguous within the frames outside $\hat{s}$ joined into one sequence, so it may straddle $\hat{s}$ in the original video. Its start is drawn uniformly at random from the positions in this sequence that leave room for the whole block. Both sides delete $m=\min(|\hat{s}|, |v|-|\hat{s}|)$ frames, where $|\cdot|$ counts frames and $|\hat{s}|$ excludes the margin. When $\hat{s}$ covers at most half of the video, $m=|\hat{s}|$ and the predicted side deletes all of $\hat{s}$. When it covers more than half, the roles swap, so the control side deletes all frames outside $\hat{s}$ and the predicted side deletes $K$ random blocks of $m$ frames inside $\hat{s}$, over which the probability is averaged instead. Each model is probed with its own prompt and inference procedure, since the question is whether its own answer depends on its own predicted span.

\paragraph{Negative values.} The random blocks are drawn only from the frames outside $\hat{s}$, so when a model answers from visual cues outside its predicted span, only the random deletions can remove those cues. Deleting $\hat{s}$ then leaves $p(v\setminus\hat{s})$ close to $p(v)$, whereas a random block occasionally covers the cues and lowers $p(v\setminus r_k)$, so that $\mathbb{E}_k[p(v\setminus r_k)] < p(v\setminus\hat{s})$ and $c < 0$. The longer $\hat{s}$ is, the larger the share of the remaining frames each random block covers, and the more often it removes these cues. This is the case for the spans shifted off the needle in \cref{sec:validate} and for the Miss items of several TGR models in \cref{tab:diag}.

\paragraph{Determinism.} Predicted spans and answers are decoded greedily, and the random blocks of each item are drawn with a fixed seed derived from its question ID, so \rg is deterministic and every value reported in this paper can be reproduced exactly.

\paragraph{Span extraction.} For the backbones, the predicted span is the envelope of all $[s,e]$ pairs inside \texttt{<think>}, from the earliest start to the latest end. For VideoChat-R1 it is the envelope of all \texttt{<timestep>} intervals, and for VideoChat-R1.5 the envelope of the intervals in the last non-empty \texttt{<glue>}. Video-Thinker gives intervals in \texttt{<time>}, of which we take the first. VisionCoach and Open-o3-Video cite point timestamps in \texttt{<t>}, and we take the window of $\pm$10\,s around the first one, clipped to the video.

\paragraph{Edge cases.} Predicted spans are clipped to the video, and an item whose span is empty after clipping counts as having no parseable span. When $\hat{s}$ covers the whole video, no frames remain for the control, and we set $c=0$ for that item.

\paragraph{Multi-turn models.} For multi-turn models, which re-watch the predicted spans in later turns, the predicted span is the union of all zoom intervals in the full-video run, the entire multi-turn loop is rerun after deletion, and each of the $K$ random controls deletes the same number of windows with matched lengths, placed uniformly at random outside the predicted span without overlapping, each with the 1\,s margin. Because the agent may request different crops after a deletion, \rg here measures whether the deleted content is needed for its final answer, rather than holding the rest of its trajectory fixed. Tool-augmented agents can also take longer trajectories with little change in their answers~\cite{li2026sponge}, so the number of crops alone does not show which content the answer uses.

\paragraph{Answer readout.} After each deletion, the model generates its full output again, and $p$ is read from this new output. We read the probabilities of the option letters at the position following the \texttt{<answer>} tag from the top-20 log-probabilities, give probability 0 to options outside them, and renormalize over the valid options. If the output contains no answer tag, we append one and read the next-token distribution. A truncated output is dropped. For a random block, this leaves the block out of the average over blocks, and for the full-video run or the deletion of $\hat{s}$, it removes the item. Probabilities are clipped at $10^{-4}$ away from 0 and 1 before taking the logit. Items whose output contains no parseable span are excluded, and both expectations in the definition of \rg are averages over the remaining items. On the Hit and Miss subsets, \rg uses the $\tau$ of all remaining items of the model.

\paragraph{Open-ended answers.} All questions evaluated in this paper are multiple choice, and \rg reads the probability of the full-video answer as a decision over a fixed option set~\cite{li2026jeva}. For open-ended questions, this probability can be replaced by a confidence score of the full-video answer under each deletion. Candidate scores include the probability of its first generated token, the likelihood of the whole answer string, and the average log-probability of its tokens, all studied for uncertainty estimation in language generation~\cite{malinin2020uncertainty, kuhn2023semantic}. We leave the comparison of these scores to future work.

\paragraph{Uncertainty.} Confidence intervals are the 2.5th and 97.5th percentiles over 10,000 bootstrap resamples of the items, with $\tau$ re-estimated on the evaluable items of every resample. A TGR model is compared with its backbone by a paired bootstrap, in which both are evaluated on the same resampled items. Each resample draws item indices from all items of the benchmark, and each model is evaluated on its own evaluable items among them, so the pairing holds even when the two models have different evaluable items.

\begin{table}[t]
\centering
\footnotesize
\setlength{\tabcolsep}{4pt}
  \begin{tabular}{lccccc}
    \toprule
     & & \multicolumn{2}{c}{\rg} & \multicolumn{2}{c}{GPU-min} \\
    \cmidrule(lr){3-4}\cmidrule(lr){5-6}
    $K$ & Calls & Backbone & \emph{SFT + GRPO} & Backbone & \emph{SFT + GRPO} \\
    \midrule
    1 & 3 & 0.620 & 1.184 & 24 & 28 \\
    2 & 4 & 0.600 & 1.154 & 32 & 37 \\
    3 & 5 & 0.598 & 1.147 & 41 & 46 \\
    5 & 7 & 0.584 & 1.127 & 57 & 64 \\
    10 & 12 & 0.581 & 1.111 & 97 & 110 \\
    \midrule
    3, three seeds & 5 & 0.577--0.598 & 1.117--1.147 & -- & -- \\
    \bottomrule
  \end{tabular}
\caption{\textbf{Effect of the number of random blocks $K$ on \bench.} \rg is computed on all items as in the All column of \cref{tab:diag}, for Qwen2.5-VL-7B (Backbone) and for \emph{SFT + GRPO} with $R_\text{IoU}$ on the same backbone. The $K{=}3$ row uses the seeds of all other results, and the last row gives the range over three seeds. Calls is the number of model calls per item, and GPU-min the cost of evaluating all 725 items, computed from the measured time per call without model loading.}
\label{tab:k_sens}
\end{table}

\paragraph{Number of random blocks.} \Cref{tab:k_sens} varies the number $K$ of random blocks on \bench for Qwen2.5-VL-7B and for \emph{SFT + GRPO} with $R_\text{IoU}$ on the same backbone. For every $K$ in the table, the \rg of the fine-tuned model is 1.91--1.93 times that of the backbone, and \rg on Hit items stays above that on Miss items for both models. \rg decreases slightly as $K$ grows, by about 3\% from $K{=}3$ to $K{=}10$, because the control probability is averaged over the $K$ blocks before the logit is taken and the logit is convex at high probability. Redrawing the random blocks at $K{=}3$ with two other seeds changes \rg by at most 0.021 for the backbone and 0.030 for the fine-tuned model, far less than the gap of 0.55 between them.

\paragraph{Cost.} \rg takes $K{+}2$ model calls per item, one on the full video, one with the predicted span deleted, and one for each random block, while accuracy alone takes one. On a single A100 GPU, with greedy decoding of at most 64 tokens, a call takes 0.67 GPU-seconds for the backbone and 0.76 for the fine-tuned model, including video decoding. At $K{=}3$, evaluating all 725 items of \bench thus takes 41 and 46 GPU-minutes, less than half the cost at $K{=}10$, while \rg differs from its $K{=}10$ value by at most 0.036. For multi-turn models, each call is a full multi-turn trajectory, so the cost grows with the number of turns.

\section{Auditing Grounded Benchmarks}
\label{app:audit}

\Cref{fig:audit_models} repeats the audit of \cref{fig:audit} with Qwen3-VL-8B and with GLM-4.1V-9B-Thinking~\cite{hong2025glm}, whose vision encoder and language model both come from outside the Qwen family. For all three models, deleting the annotated span of the four grounded benchmarks costs only slightly more accuracy than deleting a random segment of the same length, whereas deleting the needle of \bench costs far more.

\begin{figure}[t]
\centering
  \definecolor{fpInk}{HTML}{52514E}
  \definecolor{fpGrid}{HTML}{E4E3DF}
  \definecolor{auBlue}{HTML}{4A8BC4}
  \definecolor{auCoral}{HTML}{D0785C}
  \definecolor{auGreen}{HTML}{66BC95}
  \definecolor{auBand}{HTML}{F1F0EC}
\begin{lrbox}{\sidecapbox}\scriptsize\setlength{\tabcolsep}{2pt}%
  \begin{tabular}{lccccc}
    \toprule
    Dataset & Items & Full-video & Random & Annotated & $\Delta$ \\
    \midrule
    NExT-GQA (val)~\cite{nextgqa} & 573 & 76.1 & 97.3 & 90.8 & +6.6 \\
    CG-Bench~\cite{cgbench} & 379 & 38.4 & 95.4 & 89.4 & +6.0 \\
    Video-Thinker-10K~\cite{wang2025video} & 331 & 63.2 & 93.8 & 81.9 & +11.9 \\
    Seeker-173K~\cite{zeng2026video} & 838 & 83.9 & 98.3 & 83.5 & +14.8 \\
    \midrule
    \dataset & 641 & 70.7 & 94.9 & \textbf{13.6} & \textbf{+81.3} \\
    \bottomrule
  \end{tabular}
\end{lrbox}%
\setlength{\sidecapw}{\dimexpr\linewidth-\wd\sidecapbox-0.6em\relax}%
\begin{minipage}[t]{\sidecapw}\vspace{0pt}\centering
  \begin{tikzpicture}
  \begin{axis}[
    set layers, scale only axis, clip=false,
    width=0.85\linewidth, height=0.36\linewidth,
    ybar=1.2pt, bar width=6.6pt,
    xmin=0.5, xmax=5.5, xtick={1,2,3,4,5},
    xticklabels={ReliaNeedle, DeVE-QA, NExT-GQA, ReXTime, CG-Bench}, xticklabel style={font=\fontsize{5.6}{6.5}\selectfont},  %
    ymin=0, ymax=70, ytick={0,20,40,60},
    ylabel={Masking gap (pp)}, ylabel style={yshift=-3pt},
    axis x line*=bottom, axis y line*=left,
    axis line style={fpInk, line width=0.6pt},
    tick align=outside, major tick length=3pt,
    every tick/.style={fpInk, line width=0.6pt},
    xtick style={draw=none},
    tick label style={font=\fontsize{6.3}{7}\selectfont, text=fpInk},
    label style={font=\fontsize{8}{9}\selectfont},
    ymajorgrids, major grid style={fpGrid, line width=0.5pt},
    legend style={draw=none, fill=none, at={(0.98,0.98)}, anchor=north east, font=\fontsize{6.3}{7}\selectfont, text=fpInk, row sep=1pt},
    legend cell align=left,
    legend image code/.code={\fill[#1] (0pt,-2.2pt) rectangle (6pt,2.8pt);},
  ]
    \begin{pgfonlayer}{axis background}
      \fill[auBand] (axis cs:0.5,0) rectangle (axis cs:1.5,70);
    \end{pgfonlayer}
    \addplot[draw=none, fill=auBlue] coordinates {(1,62.2) (2,7.3) (3,7.4) (4,9.3) (5,9.2)};
    \addplot[draw=none, fill=auCoral] coordinates {(1,66.6) (2,5.9) (3,4.6) (4,8.8) (5,9.9)};
    \addplot[draw=none, fill=auGreen] coordinates {(1,59.2) (2,5.3) (3,4.2) (4,8.7) (5,3.4)};
    \legend{Qwen3-VL-8B, Qwen2.5-VL-7B, GLM-4.1V-9B-Thinking}
  \end{axis}
  \end{tikzpicture}
\par\vspace{6pt}\captionsetup{skip=0pt}
\caption{Accuracy after deleting a random segment minus that after deleting the annotated evidence, on the correctly answered questions of each benchmark, for Qwen3-VL-8B, Qwen2.5-VL-7B, and GLM-4.1V-9B-Thinking.}
\label{fig:audit_models}
\end{minipage}\hfill
\begin{minipage}[t]{\wd\sidecapbox}\vspace{0pt}\usebox{\sidecapbox}\par\vspace{6pt}\captionsetup{skip=0pt}
\captionof{table}{\textbf{Auditing TGR training data.} Accuracy (\%) of Qwen2.5-VL-7B on the full video (Full-video), and on its correctly answered questions after deleting a random segment (Random) or the annotated span (Annotated) of the same length. Items is the number of correctly answered questions, and $\Delta$ is Random minus Annotated.}
\label{tab:audit_train}
\end{minipage}
\end{figure}

\section{\texorpdfstring{Auditing TGR Training Data}{Auditing TGR Training Data}}
\label{app:audit_train}

We apply the audit of \cref{app:audit} to the TGR training data of the models in \cref{sec:diag} and to \dataset. NExT-GQA (val)~\cite{nextgqa} is used by VideoChat-R1, VideoChat-R1.5, LOVE-R1, and Video-o3; CG-Bench~\cite{cgbench} by LOVE-R1; Video-Thinker-10K~\cite{wang2025video} by Video-Thinker; and Seeker-173K~\cite{zeng2026video} by Video-o3. VideoZoomer trains without evidence annotations. For each dataset, we sample multiple-choice questions with annotated evidence, excluding the NExT-GQA part of Seeker-173K. Seeker-173K annotates evidence as key timestamps, so we take the source segment when available, and otherwise the union of 1\,s windows around the key timestamps, as the annotated span. We report the accuracy of Qwen2.5-VL-7B on the full video (\emph{Full-video}), and, on the questions it answers correctly, its accuracy after deleting a random segment (\emph{Random}) or the annotated span (\emph{Annotated}) of the same length. On the four TGR training sets, Qwen2.5-VL-7B keeps 81.9--90.8\% accuracy after the annotated span is deleted, only 6.0--14.8 points below a random deletion, so the annotated span is rarely the only evidence for the answer. On \dataset, accuracy drops to 13.6\%, 81.3 points below a random deletion.

\begin{figure*}[t]
\centering
\includegraphics[width=\textwidth]{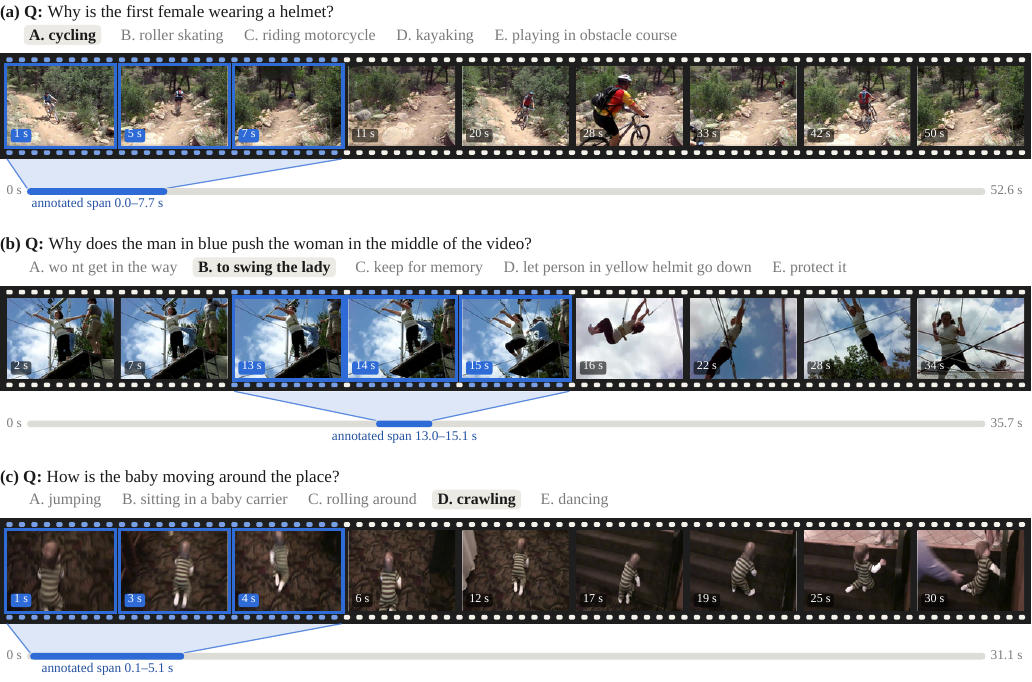}
\caption{\textbf{Annotated spans that are not needed.} Three NExT-GQA questions on which deleting the annotated span (blue) leaves the answer unchanged, since frames outside the span show the same evidence. The highlighted option is the answer.}
\label{fig:audit_examples}
\end{figure*}

\Cref{fig:audit_examples} shows three NExT-GQA questions from this audit. Their annotated spans cover only part of the evidence, and frames outside the spans show the same activity, so Qwen2.5-VL-7B, Qwen3-VL-8B, and GLM-4.1V-9B-Thinking all still answer correctly after the span is deleted, although Qwen2.5-VL-7B answers none of the three correctly without the video.

\section{\texorpdfstring{\bench}{ReliaNeedle} Construction Details}
\label{app:needle}

\paragraph{Needles and haystacks.} Needles are clips from STAR~\cite{star} and Charades~\cite{sigurdsson2016hollywood}, and haystacks are videos from Video-MME~\cite{fu2025videomme}. Each needle is inserted as one contiguous clip at a random position of its haystack. Each needle--haystack pair is verified to have no content overlap, so that the answer cannot be inferred from the surrounding video.

\paragraph{Statistics.} Haystacks last 84.4\,s on average (median 88.7\,s, 10th--90th percentile 46.2--119.0\,s, range 11.0--130.6\,s), and 71.0\% of them last between 60 and 120\,s. With the needle inserted, the full videos last 101.2\,s on average (median 104.5\,s, range 23.2--171.4\,s). Needles last 15.1\,s on average (median 12.4\,s, range 3.1--68.1\,s) and cover 16.2\% of the video on average (range 2.8--70.9\%). On \bench, frames are sampled at 1\,fps with at most 128 frames per video, so the 142 videos (19.6\%) longer than 128\,s are sampled every 1.01--1.35\,s and the others once per second. Each needle receives 14.5 frames on average (median 12, range 3--52), and every needle receives at least three frames.

\paragraph{Examples.} \Cref{fig:needle_examples} shows two questions of \bench with frames of the haystack before and after the needle.

\afterpage{%
\begin{figure*}[t]
\centering
\includegraphics[width=\textwidth]{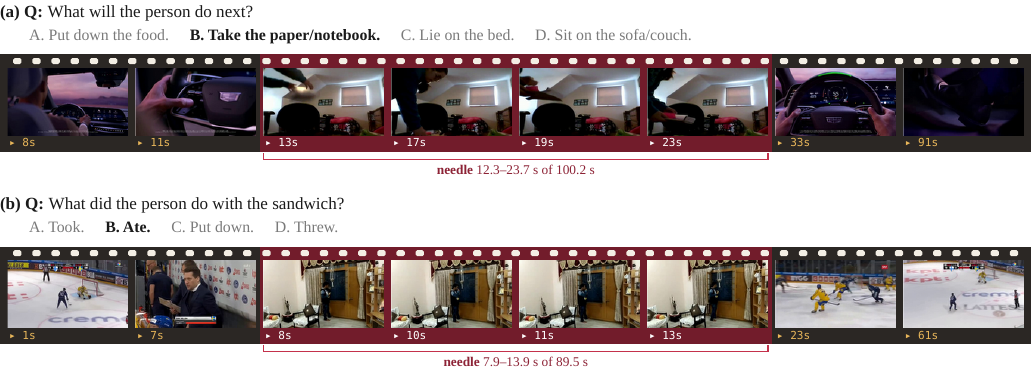}
\caption{\textbf{Examples of \bench.} Each needle (red) is a Charades clip inserted into a Video-MME video, and the question can be answered only from the needle. The bold option is the answer.}
\label{fig:needle_examples}
\end{figure*}%
}

\paragraph{Human verification.} Two annotators, one of the authors and a volunteer with a master's degree in computer science, checked a pool of 1,000 candidate questions against the criteria of \cref{sec:needle}, namely that the question is well posed and answerable only from the needle and that the needle shares no content with the haystack. Each annotator first checked 500 questions and then cross-checked the 500 checked by the other, so every question was reviewed by both, and a question was removed if either annotator rejected it. This removed 275 questions and left the 725 questions of \bench. Of the removed questions, 197 were not well posed, 32 could be answered without the needle, and 46 had a needle that shares content with the haystack. The check took three days in total.

\section{Validation without Deletion}
\label{app:human_study}

\paragraph{Judge agreement.}
We compare \rg with a measurement that deletes nothing. An independent judge, Qwen3-VL-235B~\cite{bai2025qwen3}, answers each question from only the segment a model claims, or from a random segment of the same length, and the difference in its accuracy measures how informative the predicted span is. Across nine systems (the backbone, seven of the TGR models in \cref{sec:diag}, and a model fine-tuned on certified data), this ranking agrees with that of \rg (Spearman $\rho\!=\!0.917$, permutation $p\!=\!0.0004$). Since the judge only sees where each predicted span points, this agreement rules out that the ranking is an artifact of deletion.

\begin{lrbox}{\sidecapbox}\scriptsize\setlength{\tabcolsep}{2pt}%
  \begin{tabular}{lccccc}
    \toprule
     & & & & \multicolumn{2}{c}{Rank} \\
    \cmidrule(lr){5-6}
    Model & \rg & Rating & \#Ratings & \rg & Human \\
    \midrule
    SFT on \dataset & 0.653 & 3.43 & 49 & 1 & 1 \\
    Qwen3-VL-8B & 0.235 & 3.05 & 55 & 2 & 3 \\
    VideoChat-R1 & 0.135 & 3.06 & 49 & 3 & 2 \\
    VideoChat-R1.5 & 0.063 & 2.62 & 45 & 4 & 4 \\
    Qwen2.5-VL-7B & 0.053 & 2.53 & 45 & 5 & 5 \\
    Video-Thinker & 0.025 & 2.49 & 47 & 6 & 6 \\
    \bottomrule
  \end{tabular}
\end{lrbox}%
\begin{wraptable}[13]{r}{\wd\sidecapbox}
\centering
\vspace{-\baselineskip}
\usebox{\sidecapbox}
\caption{\textbf{\rg and human ratings on Video-Oasis.} \rg is computed on the rated questions of each model, and Rating is the mean human rating from 1 to 5. SFT on \dataset fine-tunes Qwen3-VL-8B. Models are sorted by \rg.}
\label{tab:rg_user_study}
\end{wraptable}
\paragraph{Human study.} We also compare \rg with human judgments on Video-Oasis for six models. Raters see a question, the video, the model's answer, and the spans the model predicts. Without grading the answer, they find the part of the video that shows what the answer describes and rate from 1 to 5 how well the predicted spans, taken together, match that part. The items come from 94 randomly sampled videos, and the interface hides which model produced each answer. Raters rated 290 items, 45--55 per model. \Cref{tab:rg_user_study} compares the mean rating of each model with its \rg on the rated questions. The two rankings agree closely (Spearman $\rho = 0.943$, Kendall $\tau = 0.867$, exact two-sided $p = 0.017$ for both) and order 14 of the 15 model pairs the same way. The only inversion is between Qwen3-VL-8B and VideoChat-R1, whose mean ratings differ by 0.01. Fine-tuning Qwen3-VL-8B on \dataset raises \rg from 0.235 to 0.653 and the mean rating from 3.05 to 3.43, and both measures place this model first and Video-Thinker last.

\section{\texorpdfstring{\dataset}{EviCe} Pipeline Details}
\label{app:pipeline}

\subsection{Source Datasets}
\label{app:sources}
\dataset draws on twelve public datasets (\cref{tab:sources}). Six are VideoQA datasets whose question--answer pairs come without evidence spans, which \methodname annotates as described in \cref{sec:relygen}. Ego4D samples include clips that we trim from the original long videos. The other six are temporal grounding and captioning datasets that provide annotated spans with queries or captions but no questions. For these, an LLM writes a question--answer pair about the event inside each annotated span, and the resulting samples go through the same grounder agreement and evidence certification as the VideoQA samples. Matching video hashes, no video of \dataset appears in \bench or in TimeLens2-93K, the temporal grounding data added in \cref{fig:help}~(a) and (b). TimeLens-2 serves only as one of the two grounders of \methodname, and no sample of TimeLens2-93K passes through \methodname.

\subsection{\texorpdfstring{Candidate Span Generation}{Candidate Span Generation}}
\label{app:grounding}

\paragraph{Non-visual filtering.} To remove language priors, we first discard questions a model answers correctly without the video. We give Qwen3-VL-235B-A22B~\cite{bai2025qwen3} only the question and options, with the same prompt as video-based answering minus its opening ``Watch the video and'', and decode greedily. Each question is asked twice, with the original option order and with the order reversed, and is discarded only if both answers are correct, so that a positional preference such as always choosing the first option does not trigger the filter.

\paragraph{Grounding query.} For each remaining sample, an LLM combines the question and ground-truth answer into a single declarative statement describing the target visual event (the \emph{grounding query}).

\paragraph{Clip selection.} We narrow the search space by captioning each video in 30-second clips with a 25-second stride (at most 144 clips, frames sampled at 1\,fps) with Qwen3.5-122B~\cite{qwen35}. Qwen3-VL-235B-A22B~\cite{bai2025qwen3} then reads the captions and proposes up to three candidate clips that match the grounding query, each padded with 10\,s of context on both sides.

\paragraph{Grounder agreement.} Within each candidate clip, two heterogeneous visual grounding models, Keye-VL-2.0-30B-A3B~\cite{kwai2026keyevl2} and TimeLens-2 (8B)~\cite{zhu2026timelens2}, independently localize the grounding query, both from the same frames sampled at 2\,fps, with at most 192 frames and the longer side capped at 448 pixels. To guard against single-model hallucinations, we retain a localization only when the two models reach consensus with $\text{IoU} > 0.7$, taking their intersection as the candidate evidence span.

\subsection{Evidence Certification}
\label{app:certification}
\begin{figure*}[t]
\centering
\includegraphics[width=\textwidth]{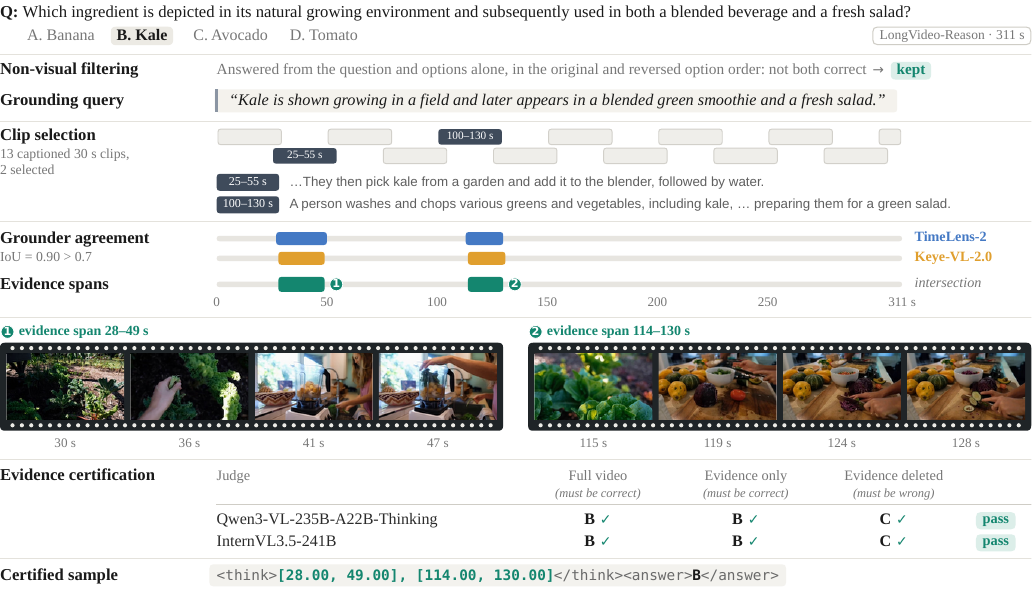}
\caption{\textbf{One sample through \methodname.} A LongVideo-Reason question passes non-visual filtering, and its grounding query selects two of the captioned clips. The two grounders agree within these clips, and their intersection gives two evidence spans. Both judges answer correctly from the full video and from the evidence alone but not after the evidence is deleted, so the sample enters \dataset.}
\label{fig:relygen_example}
\end{figure*}

Certification uses two multimodal judges from different model families, Qwen3-VL-235B-A22B-Thinking~\cite{bai2025qwen3} and InternVL3.5-241B~\cite{wang2025internvl35}, neither of which is among the models we train or evaluate. Both judges receive frames sampled at $1$\,fps, with at most $64$ frames and the longer side capped at $768$ pixels. Each judge answers every question in three settings. It must answer correctly from the full video, which ensures that the question is answerable at all, and from the evidence span alone, extended by $0.5$\,s on each side, which tests sufficiency. It must then answer incorrectly once the span, extended by $1$\,s on each side, is deleted from the video, which tests necessity. A sample is certified only if both judges pass all three settings, and the two judges run independently of each other. \Cref{fig:relygen_example} follows one sample through all steps of \methodname.

\subsection{Data Statistics}
\label{app:datastats}
\begin{figure*}[t]
\centering
\includegraphics[width=\textwidth]{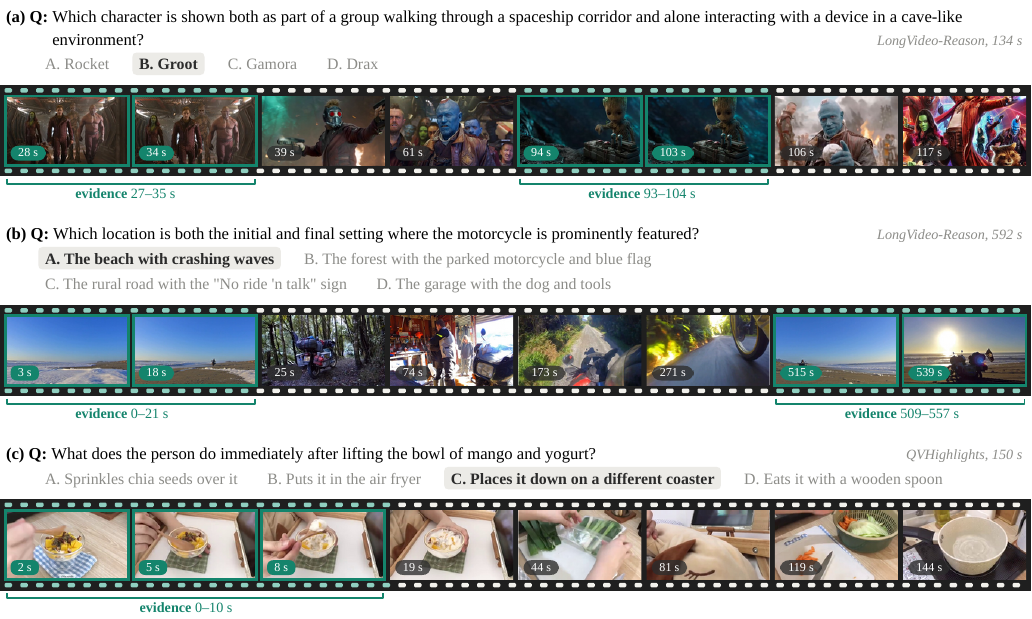}
\caption{\textbf{Certified samples of \dataset.} Green frames lie inside the evidence spans, and the highlighted option is the answer.}
\label{fig:evice_examples}
\end{figure*}

\begin{figure*}[t]
\centering
\includegraphics[width=\textwidth]{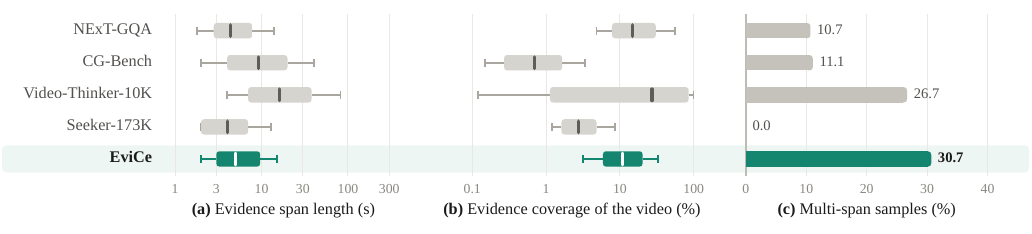}
\caption{\textbf{Evidence spans of \dataset and TGR training sets.} (a) Length of each evidence span and (b) share of the video covered by the evidence of each sample, on log scales, with boxes for the 25th--75th percentiles, whiskers for the 10th--90th, and thick ticks for the medians. (c) Share of samples with several evidence spans. Each TGR training set is represented by the 1,000 samples drawn for the audit of \cref{app:audit_train}, and \dataset by all of its samples.}
\label{fig:span_stats}
\end{figure*}

\dataset contains $20$k certified samples from twelve public datasets (\cref{tab:sources}), with videos ranging from a few seconds to over ten minutes. Unlike TGR training sets, every sample is certified to need its evidence, and a sample may carry several evidence spans when the answer combines several events. Under the audit of \cref{app:audit_train}, deleting the annotated span lowers the accuracy of Qwen2.5-VL-7B on \dataset to 13.6\%, against 81.9--90.8\% on TGR training sets. The median duration is 80.7\,s over samples and 85.9\,s over distinct videos, the mean is 94.4\,s, and 80\% of the samples last between 30.1 and 162.7\,s (\cref{fig:evice_duration}). Multi-span samples make up 30.7\% of \dataset, with 22.9\% carrying two spans and 7.7\% three or more. Their share rises with video length up to 3--5\,min, from 13.1\% for videos of at most 30\,s to 61.8\%. The median evidence span lasts 5.0\,s, similar to NExT-GQA and Seeker-173K, while multi-span samples are more common than in the four TGR training sets of \cref{app:audit_train} (\cref{fig:span_stats}). \Cref{fig:evice_examples} shows three certified samples.

\begin{figure}[t]
\centering
  \definecolor{durBlue}{HTML}{4A8BC4}
  \definecolor{durCoral}{HTML}{D0785C}
  \definecolor{fpInk}{HTML}{52514E}
  \definecolor{fpGrid}{HTML}{E4E3DF}
  \pgfplotsset{
    durbase/.style={
      /pgf/number format/assume math mode=true, yticklabel={\pgfmathprintnumber{\tick}},
      scale only axis, clip=false,
      width=0.84\linewidth, height=0.22\linewidth,
      axis x line*=bottom, axis y line*=left,
      axis line style={fpInk, line width=0.6pt},
      tick align=outside, major tick length=3pt,
      every tick/.style={fpInk, line width=0.6pt},
      tick label style={font=\fontsize{6.3}{7}\selectfont, text=fpInk},
      label style={font=\fontsize{8}{9}\selectfont},
      ylabel style={yshift=-3pt},
      ymajorgrids, major grid style={fpGrid, line width=0.5pt},
      xmin=0.4, xmax=7.6, xtick={1,2,3,4,5,6,7}, ymin=0,
    },
    durbar/.style={ybar, bar width=0.62, bar shift=0pt, draw=none, fill=durBlue},
    durbarlight/.style={durbar, fill=durBlue!35},
    durbarB/.style={durbar, fill=durCoral},
    durbarBlight/.style={durbar, fill=durCoral!35},
  }
\begin{lrbox}{\sidecapbox}\footnotesize\setlength{\tabcolsep}{4pt}%
  \begin{tabular}{lr}
    \toprule
    Source & Samples \\
    \midrule
    \multicolumn{2}{l}{\emph{VideoQA}} \\
    LLaVA-Video-178K~\cite{zhang2024llavavideo} & 12,258 \\
    E.T. Instruct~\cite{liu2024etbench} (HACS~\cite{zhao2019hacs} videos) & 789 \\
    V1-33K~\cite{wang2025nextevent} & 582 \\
    LongVideo-Reason~\cite{chen2025longvila} & 519 \\
    FineVideo~\cite{farre2024finevideo} & 267 \\
    Ego4D~\cite{grauman2022ego4d} & 159 \\
    \midrule
    \multicolumn{2}{l}{\emph{Temporal grounding and captioning}} \\
    InternVid-VTime~\cite{wang2024internvid,huang2024vtimellm} & 2,046 \\
    CosMo-Cap~\cite{wang2024cosmo} & 1,781 \\
    QVHighlights~\cite{lei2021qvhighlights} & 651 \\
    DiDeMo~\cite{hendricks2017didemo} & 446 \\
    QuerYD~\cite{oncescu2021queryd} & 388 \\
    HiREST~\cite{zala2023hirest} & 114 \\
    \bottomrule
  \end{tabular}
\end{lrbox}%
\begin{minipage}[t]{\wd\sidecapbox}\vspace{0pt}\usebox{\sidecapbox}\end{minipage}\hfill
\begin{minipage}[t]{0.5\linewidth}\vspace{0pt}\centering
  \begin{tikzpicture}
  \begin{axis}[durbase, name=durtop, ymax=35, ytick={0,10,20,30}, xticklabels={}, ylabel={Samples (\%)}]
    \addplot[durbar] coordinates {(1,9.8) (2,29.1) (3,31.7) (4,24.7)};
    \addplot[durbarlight] coordinates {(5,3.2) (6,1.1) (7,0.4)};
  \end{axis}
  \begin{axis}[durbase, at={(durtop.south west)}, anchor=north west, yshift=-10pt, ymax=65, ytick={0,20,40,60},
    xticklabels={$\le$0.5, 0.5--1, 1--2, 2--3, 3--5, 5--10, $>$10},
    xlabel={Video duration (min)}, ylabel={Multi-span (\%)}]
    \addplot[durbarB] coordinates {(1,13.1) (2,22.9) (3,36.0) (4,36.0)};
    \addplot[durbarBlight] coordinates {(5,61.8) (6,29.6) (7,30.6)};
  \end{axis}
  \end{tikzpicture}
\end{minipage}\par\vspace{6pt}
\begin{minipage}[t]{\wd\sidecapbox}\captionsetup{skip=0pt}
\captionof{table}{\textbf{Source datasets of \dataset.} The number of certified samples from each source.}
\label{tab:sources}
\end{minipage}\hfill
\begin{minipage}[t]{0.5\linewidth}\captionsetup{skip=0pt}
\caption{\textbf{Video duration in \dataset.} The upper panel gives the share of samples in each duration bin, and the lower panel the share of multi-span samples within each bin. Lighter bars mark the three bins above 3\,min, which together hold 4.7\% of the samples.}
\label{fig:evice_duration}
\end{minipage}
\end{figure}

\section{Diagnosis Details}
\label{app:diag}

\paragraph{Prompts.} \Cref{fig:prompts_a,fig:prompts_b} list the prompts of the backbones, the single-turn TGR models, and VideoZoomer. VideoChat-R1.5 uses one prompt for the first two of its three perception rounds and another for the final round, which is also used after deletion. LOVE-R1, Conan, Video-o3, and Open-o3-Video-v2 use the unchanged templates shipped with their official inference code, namely \texttt{infer.py} of LOVE-R1, \texttt{prompt\_temp.py} of Conan-Eval, the training template of Video-o3, and \texttt{video\_o3/prompts.py} of Open-o3-Video-v2.

\begin{figure*}[tp]
\scriptsize
\begin{promptbox}{Prompt 1 (Qwen2.5-VL-7B and Qwen3-VL-4B)}
\promptrole{System}
\begin{verbatim}
You are given a video and a multiple-choice question. First identify the relevant time segment in seconds
inside <think></think> as [start, end]. Then place the correct option letter inside <answer></answer>.
Example: <think>[5.2, 10.4]</think><answer>B</answer>.
\end{verbatim}
\promptrole{User}
\begin{verbatim}
Question: {question}
Options:
A. {option A}
B. {option B}
...
Return the relevant time segment in <think></think> and the correct option letter in <answer></answer>.
\end{verbatim}
\end{promptbox}
\begin{promptbox}{Prompt 2 (VisionCoach, no system prompt)}
\promptrole{User}
\begin{verbatim}
Question:{question}
A. {option A}
B. {option B}
...
Select the best answer to the multiple-choice question based on the video. You must first think about the
reasoning process in the mind and then provide the user with the answer. The reasoning process and answer are
enclosed within <think> </think> and <answer> </answer> tags, respectively. All reasoning must be grounded in
visual evidence from the video. When you mention any related object, person, or specific visual element, you
must strictly follow the following format:
`<obj>object_name</obj><box>bounding_box</box>at<t>time_in_seconds</t>s`. The reasoning process MUST NOT be
longer than 100 words. In the answer part, respond with only the letter (A, B, C, or D) of the correct option.
\end{verbatim}
\end{promptbox}
\begin{promptbox}{Prompt 3 (VideoChat-R1, no system prompt)}
\promptrole{User}
\begin{verbatim}
Answer the question: "{question}" according to the content of the video. Select the answer from :A.{option
A}, B.{option B}, ....

Output your thought process within the <think> </think> tags, including analysis with either specific
timestamps (xx.xx) or time ranges (xx.xx to xx.xx) in <timestep> </timestep> tags.

Then, provide your answer within the <answer> </answer> tags, output the corresponding letter of the option.
\end{verbatim}
\end{promptbox}
\begin{promptbox}{Prompt 4 (VideoChat-R1.5, no system prompt)}
\promptrole{User, rounds 1 and 2 of the released procedure}
\begin{verbatim}
Answer the question: "{question}" according to the content of the video. Select the answer from
:{options_inline}.

Output your think process within the  <think> </think> tags.

Then, provide your answer within the <answer> </answer> tags, output the corresponding letter of the option.
At the same time, in the <glue> </glue> tags, present the precise time period in seconds of the video clips
on which you base your answer to this question in the format of [(s1, e1), (s2, e2), ...]. For example:
<think>...</think><answer>A</answer><glue>[(5.2, 10.4)]</glue>.
\end{verbatim}
\promptrole{User, round 3 of the released procedure and after deletion}
\begin{verbatim}
Answer the question: "{question}" according to the content of the video. Select the answer from :A.{option
A}, B.{option B}, ....

Output your think process within the  <think> </think> tags.

Then, provide your answer within the <answer> </answer> tags, output the corresponding letter of the option.
\end{verbatim}
\end{promptbox}
\caption{\textbf{Prompts of the backbones and of VisionCoach, VideoChat-R1, and VideoChat-R1.5.} Placeholders are in braces, and the video precedes the text in the user turn.}
\label{fig:prompts_a}
\end{figure*}

\begin{figure*}[tp]
\scriptsize
\begin{promptbox}{Prompt 5 (Open-o3-Video)}
\promptrole{System}
\begin{verbatim}
A conversation between user and assistant. The user provides a video and asks a question, and the Assistant
solves it. The assistant MUST first think about the reasoning process in the mind and then provide the user
with the answer. The reasoning process and answer are enclosed within <think> </think> and <answer> </answer>
tags, respectively. All reasoning must be grounded in visual evidence from the video. When you mention any
related object, person, or specific visual element in the reasoning process, you must strictly follow the
following format: `<obj>object_name</obj><box>bounding_box</box>at<t>time_in_seconds</t>s`. The answer part
only requires a text response; tags like <obj>, <box>, <t> are not needed. For multiple-choice questions,
provide only the option letter in <answer> </answer>.
\end{verbatim}
\promptrole{User}
\begin{verbatim}
{question}

A. {option A}
B. {option B}
...
\end{verbatim}
\end{promptbox}
\begin{promptbox}{Prompt 6 (Video-Thinker)}
\promptrole{System}
\begin{verbatim}
You are an expert video analyst tasked with solving problems based on video content. When answering a
question about a video, you should carefully observe and analyze important visual clues from the videos to
answer. For each important segment you notice, first observe the key visual elements, then analyze their
significance using the following format: specify the time range with `<time>start_time-end_time</time>`,
describe the key visual clues with `<caption>Description of key visual clues</caption>`, and provide your
analysis about what this means with `<think>Your analysis and thoughts about this segment</think>`.
Throughout your analysis, think about the question as if you were a human pondering deeply, engaging in an
internal dialogue using natural thought expressions such as such as 'let me think', 'wait', 'Hmm', 'oh, I
see', 'let's break it down', etc, or other natural language thought expressions. After examining the key
visual clues, continue with deeper reasoning that connects your observations to the answer. Self-reflection
or verification in your reasoning process is encouraged when necessary, though if the answer is
straightforward, you may proceed directly to the conclusion. Finally, conclude by placing your final answer
in `<answer> </answer>` tags.
\end{verbatim}
\promptrole{User}
\begin{verbatim}
{question}
A. {option A}
B. {option B}
...

Please analyze the video carefully by identifying key segments and their important visual clues within
`<time> </time>`, `<caption> </caption>`, `<think> </think>` tags then conduct deep analysis and reasoning to
arrive at your answer to the question, finally provide only the single option letter (e.g., A, B, C, D, E, F
etc.) within the `<answer> </answer>` tags.Follow the format specified in the instructions.
\end{verbatim}
\end{promptbox}
\begin{promptbox}{Prompt 7 (VideoZoomer)}
\promptrole{System}
\begin{verbatim}
You are a helpful assistant. You will receive a low-frame-rate video and related questions. You can analyze
the video content to answer the question and trigger high-frame-rate inspections when finer temporal
resolution is needed. When you detect ambiguous motion/objects that require closer inspection, wrap your
request in <video_zoom></video_zoom> tags and provide the exact time segment and target frame rate in JSON
format: <video_zoom> {"segment": [start_sec, end_sec], "fps": n} </video_zoom>, it will return the video clip
at the target fps to help you better answer the question. Note that the total frames num of the request clip
cannot exceed 16 (e.g., (end_sec - start_sec) * fps <= 16) and DO NOT include <answer> tags in this round.
 Example usage: <video_zoom> {"segment": [4.0, 6.0], "fps": 2} </video_zoom>. If the initial tool response
does not provide sufficient information to answer the question, you may continue to request additional video
zoom inspections as needed, until you either (1) gather enough information to form a complete answer, or (2)
are explicitly instructed to stop using the tool. Output the thinking process within <think> </think> tags,
once you confirm your final answer, place the final answer in \boxed{} inside <answer> and </answer>.
\end{verbatim}
\promptrole{User}
\begin{verbatim}
Question: {question}
A. {option A}
B. {option B}
...
\end{verbatim}
\end{promptbox}
\caption{\textbf{Prompts of Open-o3-Video, Video-Thinker, and VideoZoomer.} Placeholders are in braces, and the video precedes the text in the user turn.}
\label{fig:prompts_b}
\end{figure*}

\paragraph{Reliance and localization.} Because \textbf{All} averages \rg over Hit and Miss items, it partly tracks how often a model's predicted span hits the evidence (R@0.3). Within the \textbf{Hit} subset, where every predicted span overlaps the evidence, \rg is unrelated to R@0.3 across the TGR models of \cref{tab:diag} (Spearman $\rho = 0.10$), so their lower reliance is not explained by localization quality.

\begin{table}[t]
\centering
\footnotesize
\setlength{\tabcolsep}{3pt}
  \begin{tabular}{lcccccccc}
    \toprule
    & \multicolumn{2}{c}{Span} & \multicolumn{3}{c}{Log-odds contrast} & \multicolumn{3}{c}{Accuracy drop} \\
    \cmidrule(lr){2-3}\cmidrule(lr){4-6}\cmidrule(lr){7-9}
    Model & (s) & (\%) & $\tau$ & $\mathbb{E}[c]$ & \rg & Rand. & Pred. & $\Delta$Acc \\
    \midrule
    Qwen2.5-VL-7B & 5.1 & 5.7 & 1.92 & 1.148 & 0.598 & $-$1.4 & 14.2 & 15.6 \\
    \midrule
    \multicolumn{9}{l}{\emph{TGR, single-turn}} \\
    VisionCoach & 19.0 & 21.3 & 5.02 & 2.352 & 0.468 & 2.4 & 17.6 & 15.2 \\
    VideoChat-R1.5 & 12.5 & 13.5 & 6.52 & 2.856 & 0.438 & 2.0 & 14.3 & 12.3 \\
    VideoChat-R1 & 8.6 & 9.3 & 6.71 & 2.514 & 0.375 & 1.6 & 11.8 & 10.2 \\
    Open-o3-Video & 16.1 & 18.7 & 4.61 & 1.250 & 0.271 & 0.5 & 11.8 & 11.3 \\
    Video-Thinker & 7.3 & 8.4 & 8.72 & $-$0.097 & $-$0.011 & 2.4 & 1.9 & $-$0.5 \\
    \midrule
    \multicolumn{9}{l}{\emph{TGR, multi-turn}} \\
    Video-o3 & 6.8 & 7.5 & 8.83 & 5.399 & 0.611 & $-$1.4 & 23.0 & 24.4 \\
    LOVE-R1 & 7.7 & 8.3 & 7.94 & 3.916 & 0.493 & 0.5 & 16.2 & 15.7 \\
    VideoZoomer & 1.7 & 1.9 & 7.57 & 1.477 & 0.195 & $-$1.3 & 5.0 & 6.3 \\
    \bottomrule
  \end{tabular}
\caption{\textbf{Span length, confidence scale, and accuracy on \bench.} Span is the mean length of the predicted span, in seconds and as a percentage of the video, where the span of a multi-turn model is the union of its zoom intervals. $\tau$ is the mean absolute full-video log-odds and $\mathbb{E}[c]$ the mean deletion contrast before normalization, so that \rg${}=\mathbb{E}[c]/\tau$, the All column of \cref{tab:diag}. Rand. and Pred. are the drops in accuracy (points) over all evaluated items after deleting a random segment or the predicted span, and $\Delta$Acc is Pred. minus Rand.}
\label{tab:len_tau}
\end{table}
\paragraph{Span length, confidence scale, and accuracy.} \rg divides the mean deletion contrast $\mathbb{E}[c]$ by each model's confidence scale $\tau$, and models differ in how long their predicted spans are. \Cref{tab:len_tau} reports both factors for the models of \cref{tab:diag}, together with the drop in accuracy after deleting a random segment or the predicted span. \rg is unrelated to the mean span length (Spearman $\rho = -0.07$). The TGR models are far more confident than the backbone ($\tau = 4.6$--$8.8$ versus $1.92$), and log-odds changes grow with confidence, so their unnormalized $\mathbb{E}[c]$ exceeds that of the backbone even when deletion hurts them less. Normalizing by $\tau$ removes this scale effect, and among the TGR models, whose confidence scales are closer, it barely changes the ranking ($\rho = 0.90$ between $\mathbb{E}[c]$ and \rg). The accuracy contrast, which does not depend on this scale, agrees with \rg (Spearman $\rho = 0.97$). Deleting a random segment changes accuracy by at most 2.4 points, while deleting the predicted span costs up to 23.0 points, and every single-turn TGR model again falls below the backbone.

\paragraph{Confidence intervals.}\phantomsection\label{app:ci} \Cref{tab:ci} gives bootstrap confidence intervals for \cref{tab:diag} and compares each TGR model with its own backbone on the same resampled items. No TGR model has a significantly higher \rg than its backbone. Video-o3 is the only one with a positive difference ($+0.013$, $p{=}0.40$), and for eight of the other nine, the 95\% interval of the difference lies entirely below zero.

\begin{table*}[t]
  \centering
  \footnotesize
  \setlength{\tabcolsep}{2.6pt}  %
  \def\cint#1#2{\unskip\,{\scriptsize[#1, #2]}}
  \begin{tabular}{@{}lcccccc@{}}  %
    \toprule
     & & \multicolumn{3}{c}{\rg} & \multicolumn{2}{c}{vs.\ own backbone} \\
    \cmidrule(lr){3-5}\cmidrule(lr){6-7}
    Model & Items & All & Hit & Miss & $\Delta$ & $p$ \\
    \midrule
    \multicolumn{7}{l}{\emph{General-purpose}} \\
    Qwen2.5-VL-7B & 713 & 0.598 \cint{0.523}{0.672} & 0.945 \cint{0.811}{1.083} & 0.302 \cint{0.225}{0.383} & -- & -- \\
    Qwen3-VL-4B$^\dagger$ & 519 & 0.711 \cint{0.640}{0.780} & 1.026 \cint{0.929}{1.123} & 0.320 \cint{0.211}{0.430} & -- & -- \\
    \midrule
    \multicolumn{7}{l}{\emph{TGR, single-turn}} \\
    VisionCoach & 703 & 0.468 \cint{0.388}{0.548} & 0.763 \cint{0.660}{0.865} & $-$0.019 \cint{$-$0.127}{0.092} & $-$0.130 \cint{$-$0.227}{$-$0.032} & 0.996 \\
    VideoChat-R1.5 & 703 & 0.438 \cint{0.366}{0.511} & 0.863 \cint{0.748}{0.975} & 0.060 \cint{$-$0.011}{0.133} & $-$0.160 \cint{$-$0.254}{$-$0.066} & 0.999 \\
    VideoChat-R1 & 710 & 0.375 \cint{0.300}{0.451} & 0.709 \cint{0.590}{0.832} & 0.081 \cint{$-$0.002}{0.166} & $-$0.223 \cint{$-$0.321}{$-$0.123} & $>$0.999 \\
    Open-o3-Video & 680 & 0.271 \cint{0.191}{0.350} & 0.764 \cint{0.625}{0.902} & $-$0.100 \cint{$-$0.174}{$-$0.026} & $-$0.327 \cint{$-$0.427}{$-$0.227} & $>$0.999 \\
    Video-Thinker & 572 & $-$0.011 \cint{$-$0.067}{0.045} & 0.384 \cint{0.194}{0.577} & $-$0.095 \cint{$-$0.147}{$-$0.043} & $-$0.609 \cint{$-$0.704}{$-$0.514} & $>$0.999 \\
    \midrule
    \multicolumn{7}{l}{\emph{TGR, multi-turn}} \\
    Video-o3 & 725 & 0.611 \cint{0.532}{0.689} & 0.700 \cint{0.595}{0.802} & 0.448 \cint{0.330}{0.572} & $+$0.013 \cint{$-$0.088}{$+$0.114} & 0.395 \\
    LOVE-R1 & 684 & 0.493 \cint{0.413}{0.573} & 0.628 \cint{0.527}{0.729} & 0.218 \cint{0.096}{0.345} & $-$0.105 \cint{$-$0.210}{$+$0.001} & 0.974 \\
    Conan & 725 & 0.240 \cint{0.169}{0.314} & 0.489 \cint{0.328}{0.655} & 0.180 \cint{0.101}{0.262} & $-$0.358 \cint{$-$0.453}{$-$0.261} & $>$0.999 \\
    VideoZoomer & 715 & 0.195 \cint{0.141}{0.249} & 0.693 \cint{0.414}{0.980} & 0.153 \cint{0.100}{0.206} & $-$0.403 \cint{$-$0.493}{$-$0.315} & $>$0.999 \\
    Open-o3-Video-v2$^\dagger$ & 705 & 0.597 \cint{0.515}{0.676} & 0.651 \cint{0.556}{0.741} & 0.413 \cint{0.249}{0.578} & $-$0.114 \cint{$-$0.220}{$-$0.010} & 0.983 \\
    \bottomrule
  \end{tabular}
  \caption{\textbf{Bootstrap confidence intervals for \cref{tab:diag}.} Items is the number of items with a parseable span, and brackets give 95\% percentile intervals over 10,000 resamples of the 725 items. Hit and Miss use the $\tau$ of all items of the model. $\Delta$ is the \rg of a TGR model minus that of its own backbone on the same resampled items, and $p$ is the one-sided $p$-value for $\Delta > 0$. $^\dagger$ marks Qwen3-VL-4B and the TGR model built on it; all other TGR models are built on Qwen2.5-VL-7B.}
  \label{tab:ci}
\end{table*}

\paragraph{Robustness to the confidence scale.} The TGR models are far more confident than their backbones, and several of them place most items at the clipping bound (\cref{tab:scale}). Two checks show that the diagnosis does not hinge on this scale. First, \rg barely changes with the clipping threshold even though $\tau$ does. For Video-o3, $\tau$ is 6.71, 8.83, and 12.71 at thresholds of $10^{-3}$, $10^{-4}$, and $10^{-6}$, while \rg is 0.607, 0.611, and 0.621. At all three thresholds, nine of the ten TGR models have a lower \rg than their backbone, and none is significantly higher. Second, the flip gap, which counts how often the answer changes rather than how much the confidence drops, does not depend on the confidence scale. Under it, eight of the ten TGR models fall below their backbone, five of them significantly, and Video-o3 is the only one significantly above it, in line with the accuracy contrast of \cref{tab:len_tau}. Open-o3-Video and Conan saturate on at most 10\% of their items, yet both are significantly below their backbone under both measures.

\begin{table}[t]
\centering
\footnotesize
\setlength{\tabcolsep}{3pt}
  \begin{tabular}{lccc}
    \toprule
    Model & Sat. (\%) & Flip gap (pp) & $\Delta$ vs.\ backbone \\
    \midrule
    \multicolumn{4}{l}{\emph{General-purpose}} \\
    Qwen2.5-VL-7B & 1.4 & 24.4 & -- \\
    Qwen3-VL-4B$^\dagger$ & 24.3 & 45.3 & -- \\
    \midrule
    \multicolumn{4}{l}{\emph{TGR, single-turn}} \\
    VisionCoach & 8.5 & 22.3 & $-$2.1 {\scriptsize[$-$6.9, $+$2.8]} \\
    VideoChat-R1.5 & 33 & 24.3 & $-$0.0 {\scriptsize[$-$5.0, $+$4.6]} \\
    VideoChat-R1 & 24 & 20.1 & $-$4.2 {\scriptsize[$-$9.0, $+$0.6]} \\
    Open-o3-Video & 5.9 & 12.4 & $-$12.0 {\scriptsize[$-$16.8, $-$7.1]} \\
    Video-Thinker & 75 & 0.4 & $-$23.9 {\scriptsize[$-$28.4, $-$19.6]} \\
    \midrule
    \multicolumn{4}{l}{\emph{TGR, multi-turn}} \\
    Video-o3 & 90 & 30.1 & $+$5.7 {\scriptsize[$+$0.7, $+$10.8]} \\
    LOVE-R1 & 32 & 25.2 & $+$0.8 {\scriptsize[$-$4.4, $+$6.1]} \\
    Conan & 10 & 11.0 & $-$13.3 {\scriptsize[$-$18.2, $-$8.4]} \\
    VideoZoomer & 57 & 10.1 & $-$14.2 {\scriptsize[$-$18.7, $-$9.6]} \\
    Open-o3-Video-v2$^\dagger$ & 90 & 30.4 & $-$15.0 {\scriptsize[$-$20.6, $-$9.3]} \\
    \bottomrule
  \end{tabular}
\caption{\textbf{Confidence saturation and flip rates on \bench.} Sat. is the share of evaluable items whose full-video probability reaches the clipping bound $1-10^{-4}$. Flip gap is the share of items whose answer changes after deleting the predicted span minus that after deleting a random segment, with abstentions counted as changes in both. $\Delta$ is the flip gap minus that of the model's own backbone, with a 95\% paired bootstrap interval. $^\dagger$ marks Qwen3-VL-4B and the TGR model built on it; all other TGR models are built on Qwen2.5-VL-7B.}
\label{tab:scale}
\end{table}

\begin{table*}[!t]
  \centering
  \small
  \setlength{\tabcolsep}{4.5pt}
  \begin{tabular}{llccccccccc}
    \toprule
    & & \multicolumn{4}{c}{Qwen3-VL-8B} & & \multicolumn{4}{c}{Qwen2.5-VL-7B} \\
    \cmidrule(lr){3-6}\cmidrule(lr){8-11}
    & & \multicolumn{2}{c}{\bench} & \multicolumn{2}{c}{Video-Oasis} & & \multicolumn{2}{c}{\bench} & \multicolumn{2}{c}{Video-Oasis} \\
    \cmidrule(lr){3-4}\cmidrule(lr){5-6}\cmidrule(lr){8-9}\cmidrule(lr){10-11}
    Start & Reward & Acc. & \rg & Acc. & \rg & & Acc. & \rg & Acc. & \rg \\
    \midrule
    \multicolumn{11}{l}{\emph{Single-turn}} \\
    \multicolumn{2}{l}{Backbone} & 60.0 & 0.613 & 34.46 & 0.213 & & 57.2 & 0.598 & 29.79 & 0.087 \\
    \multicolumn{2}{l}{SFT on \dataset} & 62.1 & 1.237 & 35.88 & 0.690 & & 59.4 & 1.066 & 34.20 & 0.411 \\
    \multicolumn{2}{l}{SFT on 1/3 of \dataset} & 63.3 & 1.145 & 34.58 & 0.608 & & 57.2 & 1.066 & 33.0 & 0.35 \\
    \midrule
    Backbone & $R_\text{ans}$ & 67.6 & 0.686 & 37.14 & 0.205 & & 64.1 & 0.166 & 32.56 & 0.048 \\
    Backbone & $R_\text{ans+IoU}$ & 68.4 & 0.917 & 36.68 & 0.363 & & 62.9 & 0.696 & 32.48 & 0.161 \\
    Backbone & $R_\text{IoU}$ & 67.7 & 0.872 & 36.43 & 0.361 & & 64.0 & 0.543 & 30.55 & 0.148 \\
    \midrule
    SFT & $R_\text{ans}$ & 64.3 & 1.095 & 35.96 & 0.473 & & 61.2 & 0.862 & 34.81 & 0.275 \\
    SFT & $R_\text{ans+IoU}$ & 64.6 & 1.032 & 36.18 & 0.551 & & 61.5 & 0.964 & 34.91 & 0.309 \\
    SFT & $R_\text{IoU}$ & 64.0 & \textbf{1.225} & 36.08 & \textbf{0.667} & & 60.8 & \textbf{1.147} & 34.70 & \textbf{0.437} \\
    \midrule
    \multicolumn{11}{l}{\emph{Multi-turn (Video-o3-style agent)}} \\
    \multicolumn{2}{l}{SFT on \dataset} & 67.2 & 1.172 & 36.09 & 0.565 &  & 65.7 & 1.000 & 30.17 & 0.441 \\
    Backbone & $R_\text{ans}$ & 69.8 & 0.928 & 38.07 & 0.293 &  & 67.4 & 0.919 & 30.26 & 0.167 \\
    SFT & $R_\text{ans}$ & 67.6 & 1.098 & 36.35 & 0.530 &  & 67.2 & 0.999 & 32.82 & 0.385 \\
    SFT & $R_\text{IoU}$ & 68.8 & 1.170 & 36.74 & 0.583 &  & 65.7 & 1.044 & 32.10 & 0.483 \\
    \bottomrule
  \end{tabular}
  \caption{Effect of the training recipe, with \dataset fixed. GRPO starts either from the backbone (\emph{GRPO w/o cold-start}) or from a cold-start SFT model (\emph{SFT + GRPO}), and rewards the answer ($R_\text{ans}$), the answer and the temporal IoU of the predicted span ($R_\text{ans+IoU}$), or the IoU alone ($R_\text{IoU}$). Bold marks the highest \rg among single-turn GRPO runs. The rows of SFT on \dataset are trained on the full \dataset. \emph{SFT + GRPO} instead runs one epoch of SFT on one third of \dataset and GRPO on the remaining, video-disjoint two thirds, so that no question is seen twice. SFT on 1/3 of \dataset is this starting point, and its Video-Oasis values on Qwen2.5-VL-7B are those at step 0 of \cref{fig:plane}b. The multi-turn block reports Video-o3-style agents (\cref{app:agent}) that crop the predicted span in later turns while keeping the full video in context.}
  \label{tab:objective}
\end{table*}

\section{Training Details}
\label{app:training}

\paragraph{Setup.} We use Qwen2.5-VL-7B and Qwen3-VL-8B~\cite{bai2025qwen3} as backbones. All runs within a comparison share the training questions, their order, the number of steps, and the hyperparameters, and differ only in the factor under study. We also continue training an existing TGR model on \dataset to test whether the data transfers. Since \rg needs no evidence annotations, we report accuracy and \rg mainly on the real videos of Video-Oasis, and give the corresponding results on \bench in \cref{app:extra}. On Video-Oasis, we evaluate on the 11,022 questions whose videos are available.

\paragraph{SFT.} We fine-tune all parameters of the language model while freezing the vision encoder and the projector, with a learning rate of $10^{-5}$, a cosine schedule with $5\%$ warmup, a global batch size of $16$, bf16 precision, and DeepSpeed ZeRO-3. Videos are sampled at $1$\,fps with at most $128$ frames and at most $602{,}112$ pixels per frame, and sequences are truncated at $51{,}200$ tokens. Both backbones use the same settings, and the cold-start model that GRPO starts from trains for $292$ steps.

\paragraph{Rewards.} In the single-turn format, all three rewards share a format gate $f \in \{0,1\}$, which requires a \texttt{<think>} block that contains at least one valid span $[s, e]$ and nothing else, followed by an \texttt{<answer>} block. With $\mathrm{Acc} \in \{0,1\}$ the correctness of the answer and $\mathrm{IoU}$ the temporal IoU between the union of the predicted spans and the ground-truth evidence span, the rewards are
\begin{align*}
R_{\text{ans}} &= f\,(\mathrm{Acc} + 0.1), \\
R_{\text{ans+IoU}} &= f\,(\mathrm{Acc} + 0.1 + 0.5\,\mathrm{IoU}), \\
R_{\text{IoU}} &= f\,(0.1 + 0.5\,\mathrm{IoU}).
\end{align*}
$R_{\text{IoU}}$ does not reward a correct answer at all, and since the three rewards share the same format gate, their differences come only from the answer and IoU terms.

\paragraph{GRPO.} Single-turn GRPO runs use a learning rate of $10^{-6}$, 16 questions with 8 rollouts each per step, a sampling temperature of 1.0, a clipping ratio of 0.2, and no KL penalty. \emph{GRPO w/o cold-start} and \emph{SFT + GRPO} on Qwen3-VL-8B share the same 5,168 questions, in the same order, for 323 steps. On Qwen2.5-VL-7B, GRPO runs for 408 steps. On both backbones and for both architectures, the SFT model that GRPO starts from trains for one third of an epoch on \dataset after removing every video used by the GRPO questions, so that both starting points meet the GRPO questions for the first time. The SFT rows of \cref{tab:data,tab:objective} are trained on the full set. Rollouts are generated with vLLM, with prompts of up to $28{,}672$ tokens and responses of up to $256$ tokens, and frames are sampled as in evaluation, at $1$\,fps with at most $128$ frames. When temporal grounding data is added in \cref{fig:help}~(a) and (b), every run keeps the same number of GRPO steps.

\section{\texorpdfstring{Multi-turn Agent}{Multi-turn Agent}}
\label{app:agent}

\paragraph{Video-o3-style agent.} The agent of \cref{sec:arch} is built on Qwen2.5-VL-7B and Qwen3-VL-8B and follows Video-o3~\cite{zeng2026video}. Its first turn always sees the full video. In each later turn it sees every crop requested so far, each labeled with its time range, and either crops again or answers. Later turns also keep the full video and the earlier reasoning, as in the original Video-o3. For SFT, each sample of \dataset is rewritten as a tool-use dialogue. The first turn reasons over the full video and crops the annotated evidence span, slightly widened, and the next turn sees the crop, states the evidence, and answers. Samples with several evidence spans receive several crops.

\paragraph{Turns and frames.} The agent runs for at most six turns, and a trajectory holds at most twelve video inputs, namely the full-video overview and up to eleven crops. The overview samples the full video at $2$\,fps with at most $768$ frames and fits within $16{,}384$ tokens. To crop, the model emits a JSON request inside \texttt{<grounding>} that specifies a temporal segment $[s, e]$ and a sampling strategy, and frames are then sampled within that segment. Each turn generates at most $8{,}192$ tokens, the context is capped at $32{,}768$ tokens, and training and evaluation use the same limits.

\paragraph{RL.} Agent GRPO uses a learning rate of $10^{-6}$, $32$ questions with $16$ rollouts each per step, no KL penalty, and one epoch over the question pool. Each trajectory is split into one training sample per turn, and all turns share the advantage of the trajectory.

\paragraph{Rewards.} For the agent, $R_{\text{ans}}$ is the official Video-o3 reward $\mathrm{Acc}\,(1 + b) + r_{\text{fmt}}$. Here $r_{\text{fmt}}$ deducts $1/T$ for each of the $T$ turns that breaks the required format, and the bonus $b = (\mathrm{IoU} + 1)/2$ is given only when the answer is correct, the format is intact, and at least one crop is requested, with $\mathrm{IoU}$ computed between the union of the requested segments and the evidence span. $R_{\text{IoU}}$ instead gives $0.1 + 0.5\,\mathrm{IoU}$ to well-formed trajectories and does not reward the answer.

\makeatletter
\setlength{\@fptop}{0pt}
\setlength{\@dblfptop}{0pt}
\renewcommand{\dbltopfraction}{0.85}
\makeatother

\section{Additional Results}
\label{app:extra}

\paragraph{Certifying TGR training data.} \cref{tab:pilot_full} extends \cref{tab:pilot} with results on Video-Oasis, where again only the certified samples raise \rg above the backbone.

\begin{table}[!ht]
\centering
\begin{lrbox}{\sidecapbox}\footnotesize\setlength{\tabcolsep}{3pt}%
  \begin{tabular}{lcccccc}
    \toprule
    & & \multicolumn{3}{c}{\bench} & \multicolumn{2}{c}{Video-Oasis} \\
    \cmidrule(lr){3-5}\cmidrule(lr){6-7}
    Model & Size & Acc. & R@0.3 & \rg & Acc. & \rg \\
    \midrule
    Qwen2.5-VL-7B~\cite{bai2025qwen25vltechnicalreport} & -- & 57.2 & 46.0 & 0.60 & 29.8 & 0.09 \\
    \quad + TGR-Mix & 20k & 58.2 & 61.3 & 0.44 & 30.2 & 0.08 \\
    \quad + TGR-Mix, random & 2k & 57.0 & 46.6 & 0.48 & 28.7 & 0.07 \\
    \quad + TGR-Mix, certified & 2k & 57.2 & 46.8 & \textbf{0.75} & 30.0 & \textbf{0.28} \\
    \bottomrule
  \end{tabular}
\end{lrbox}%
\begin{lrbox}{\sidecapboxb}\footnotesize\setlength{\tabcolsep}{3pt}%
  \begin{tabular}{llcccc}
    \toprule
     &  & \multicolumn{2}{c}{\bench} & \multicolumn{2}{c}{Video-Oasis} \\
    \cmidrule(lr){3-4}\cmidrule(lr){5-6}
    Start & Reward & Acc. & \rg & Acc. & \rg \\
    \midrule
    \multicolumn{2}{l}{Backbone} & 53.9 & 0.656 & 32.28 & 0.160 \\
    \multicolumn{2}{l}{SFT on \dataset} & 54.6 & 1.370 & 32.45 & 0.732 \\
    \multicolumn{2}{l}{SFT on 1/3 of \dataset} & 57.5 & 1.331 & 34.09 & 0.711 \\
    \midrule
    Backbone & $R_\text{ans}$ & 62.2 & 0.845 & 35.67 & 0.240 \\
    SFT & $R_\text{ans}$ & 60.7 & 0.797 & 32.87 & 0.318 \\
    SFT & $R_\text{IoU}$ & 58.9 & \textbf{1.090} & 33.99 & \textbf{0.718} \\
    \bottomrule
  \end{tabular}
\end{lrbox}%
\begin{minipage}[t]{\wd\sidecapbox}\vspace{0pt}\usebox{\sidecapbox}\par\vspace{6pt}\captionsetup{skip=0pt}
\caption{\textbf{Certifying TGR training data on both benchmarks.} Full version of \cref{tab:pilot}, adding results on Video-Oasis.}
\label{tab:pilot_full}
\end{minipage}\hfill
\begin{minipage}[t]{\wd\sidecapboxb}\vspace{0pt}\usebox{\sidecapboxb}\par\vspace{6pt}\captionsetup{skip=0pt}
\caption{\textbf{Training recipes on GLM-4.1V-9B-Thinking.} A subset of the single-turn recipes of \cref{tab:objective}, with columns and bold as there. GRPO in the SFT rows starts from SFT on 1/3 of \dataset.}
\label{tab:glm}
\end{minipage}
\end{table}

\begin{figure}[!ht]
\centering
  \definecolor{fpGray}{HTML}{8A8984}
  \definecolor{fpGreen}{HTML}{5AAF7D}
  \definecolor{fpOrange}{HTML}{CB7A63}
  \definecolor{fpBlue}{HTML}{599CC0}
  \definecolor{fpInk}{HTML}{52514E}
  \definecolor{fpGrid}{HTML}{E4E3DF}
  \pgfplotsset{
    fpbase/.style={
      /pgf/number format/assume math mode=true, xticklabel={\pgfmathprintnumber{\tick}}, yticklabel={\pgfmathprintnumber{\tick}},
      scale only axis, clip=false,
      axis x line*=bottom, axis y line*=left,
      axis line style={fpInk, line width=0.6pt},
      tick align=outside, major tick length=3pt,
      every tick/.style={fpInk, line width=0.6pt},
      tick label style={font=\fontsize{6.3}{7}\selectfont, text=fpInk},
      label style={font=\fontsize{8}{9}\selectfont},
      xlabel style={yshift=2pt}, ylabel style={yshift=-3pt},
      grid=major, major grid style={fpGrid, line width=0.5pt},
    },
    fpdot/.style={only marks, mark=*, mark size=2.55pt, mark options={fill=#1, draw=white, line width=0.7pt}},
    fptri/.style={only marks, mark=triangle*, mark size=3.3pt, mark options={fill=#1, draw=white, line width=0.7pt}},
  }
  \def\fplab#1#2#3#4{\node[font=\fontsize{6}{7}\selectfont, text=fpInk, anchor=base west, xshift=#3pt, yshift=#4pt, inner sep=0pt] at (axis cs:#1) {#2};}
\begin{minipage}{0.5\linewidth}\centering
  \begin{tikzpicture}
  \begin{axis}[fpbase, width=0.84\linewidth, height=0.5\linewidth,
    xmin=0.15, xmax=0.75, ymin=34.0, ymax=38.5,
    xtick={0.2,0.3,0.4,0.5,0.6,0.7}, ytick={34,35,36,37,38},
    xticklabel style={/pgf/number format/.cd, fixed, fixed zerofill, precision=1},
    xlabel={\rg}, ylabel={Accuracy (\%)},
    legend columns=2, legend cell align=left,
    legend style={draw=none, at={(0.5,1.03)}, anchor=south, font=\fontsize{6.6}{8}\selectfont, /tikz/every even column/.append style={column sep=8pt}}]
    \addlegendimage{legend image code/.code={\fill[fpOrange] (0cm,-0.05cm) rectangle (0.28cm,0.08cm);}}\addlegendentry{GRPO w/o cold-start}
    \addlegendimage{legend image code/.code={\fill[fpBlue] (0cm,-0.05cm) rectangle (0.28cm,0.08cm);}}\addlegendentry{SFT + GRPO}
    \addplot[fpdot=fpGray] coordinates {(0.213,34.46)};
    \addplot[fpdot=fpGreen] coordinates {(0.690,35.88)};
    \addplot[fpdot=fpOrange] coordinates {(0.205,37.14) (0.363,36.68) (0.361,36.43)};
    \addplot[fpdot=fpBlue] coordinates {(0.473,35.96) (0.551,36.18) (0.667,36.08)};
    \addplot[fptri=fpOrange] coordinates {(0.293,38.07)};
    \addplot[fptri=fpBlue] coordinates {(0.530,36.35) (0.583,36.74)};
    \fplab{0.213,34.46}{Backbone}{5}{-2}
    \fplab{0.690,35.88}{\textcolor{fpGreen}{SFT}}{3}{-8}
    \fplab{0.205,37.14}{ans}{4}{-2}
    \fplab{0.363,36.68}{ans+IoU}{-6}{4}
    \fplab{0.361,36.43}{IoU}{4}{-3}
    \fplab{0.473,35.96}{ans}{-4}{-9}
    \fplab{0.551,36.18}{ans+IoU}{3}{-9}
    \fplab{0.667,36.08}{IoU}{-4}{4}
    \fplab{0.583,36.74}{IoU}{4}{-2}
    \fplab{0.530,36.35}{ans}{-14}{-1}
    \fplab{0.293,38.07}{ans}{4}{-2}
    \draw[fpGray!60, line width=0.5pt, fill=white, rounded corners=1pt] (axis cs:0.565,34.17) rectangle (axis cs:0.718,34.93);
    \addplot[only marks, mark=*, mark size=2.3pt, mark options={fill=fpInk, draw=white}, forget plot] coordinates {(0.585,34.72)};
    \addplot[only marks, mark=triangle*, mark size=3pt, mark options={fill=fpInk, draw=white}, forget plot] coordinates {(0.585,34.38)};
    \node[anchor=west, font=\fontsize{6.3}{7}\selectfont, text=fpInk, inner sep=0pt] at (axis cs:0.603,34.72) {single-turn};
    \node[anchor=west, font=\fontsize{6.3}{7}\selectfont, text=fpInk, inner sep=0pt] at (axis cs:0.603,34.38) {multi-turn};
  \end{axis}
  \end{tikzpicture}
\caption{Accuracy and \rg on Video-Oasis for Qwen3-VL-8B, drawn as in \cref{fig:plane}a. As on Qwen2.5-VL-7B, \emph{GRPO w/o cold-start} leaves \rg far below that of \emph{SFT}, and \emph{SFT + GRPO} keeps it only with $R_{\text{IoU}}$.}
\label{fig:plane_qwen3}
\end{minipage}
\end{figure}
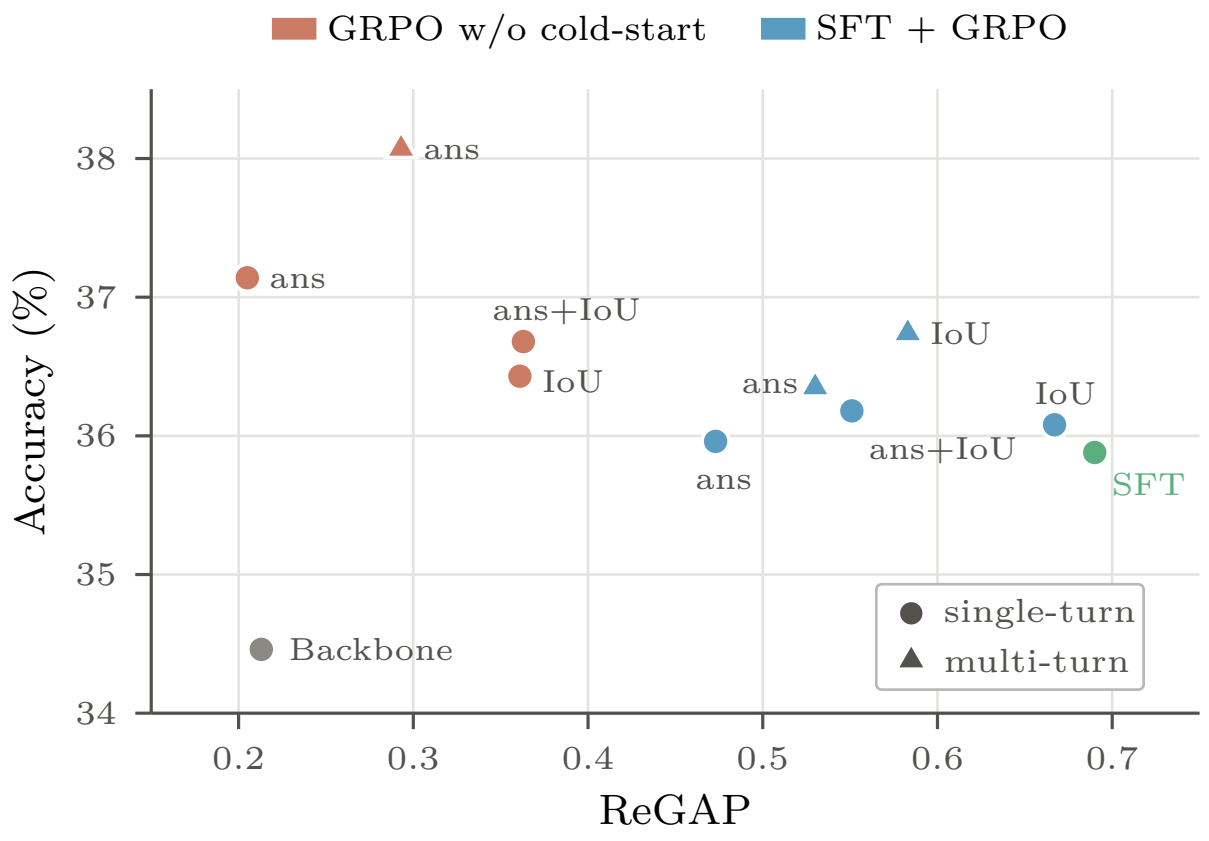

\Cref{tab:objective} lists the full results behind Fig.~\ref{fig:plane}a and~\ref{fig:plane_qwen3} on both benchmarks. Relative to its starting point, SFT on 1/3 of \dataset, single-turn \emph{SFT + GRPO} with $R_\text{IoU}$ raises both accuracy and \rg on both Qwen backbones and both benchmarks, whereas $R_\text{ans}$ and $R_\text{ans+IoU}$ raise accuracy but lower \rg.

\paragraph{Another model family.} \Cref{tab:glm} repeats part of the single-turn recipes on GLM-4.1V-9B-Thinking~\cite{hong2025glm}, a model outside the Qwen family, with the same data, frame sampling, and number of GRPO steps as for the Qwen backbones. On both benchmarks, SFT on \dataset at least doubles the \rg of the backbone, while \emph{GRPO w/o cold-start} gives the highest accuracy but leaves \rg far below that of \emph{SFT}. Starting from SFT on 1/3 of \dataset, with \rg of 1.331 on \bench and 0.711 on Video-Oasis, GRPO with $R_\text{ans}$ lowers \rg to 0.797 and 0.318, whereas $R_\text{IoU}$ retains most of it, at 1.090 and 0.718.

\section{Limitations}
\label{app:limitations}

\paragraph{Models and training.} We train models with 7 to 9B parameters. Most of them come from the Qwen family, and on GLM-4.1V-9B-Thinking we run only part of the single-turn recipes. All multi-turn results use the Video-o3-style agent. Larger models and other agent designs remain untested.

\paragraph{Scale of \dataset.} \dataset contains 20k certified samples, a size limited by our compute, since \methodname runs several large LLMs and multimodal LLMs on every candidate. How accuracy and \rg change with a larger \dataset remains untested.

\section{Broader Impact}
\label{app:impact}

Users may check a TGR model's answer by inspecting its short predicted span instead of the whole video, and a span the answer does not rely on makes it look better supported than it is. World-action models show a similar gap, where actions can shift while the imagined future stays largely unchanged~\cite{li2026badwam}. \rg and \bench test whether the shown span is the evidence a model uses, and training on \dataset with our recipe yields models whose answers follow their spans. As a model can also rely on a wrong span, \rg should be read together with accuracy and grounding. \bench and \dataset inherit the content and biases of their public source datasets, including videos of people, and should be used under their licenses.

\end{document}